\documentclass[11pt,letterpaper]{mystyle}
\usepackage{nicefrac}
\usepackage{subcaption}
\usepackage{pgfplots}
\usepackage{multirow}

\newcommand{\themename}{red}     %

\ifthenelse{\equal{\themename}{red}}{
  \newcommand{\thelogo}{}
  \definecolor{accent_primary}{HTML}{5E3545}   %
  \definecolor{accent_cite}{HTML}{52423A}      %
  \definecolor{bg_title}{HTML}{F6F4F4}         %
  \definecolor{bg_sepia}{HTML}{F7F5F5}         %
  \definecolor{bordercolor}{HTML}{d9cec8}      %

  \definecolor{perf0}{HTML}{ffffff}
  \definecolor{perf10}{HTML}{fbf5f7}
  \definecolor{perf20}{HTML}{f5ebee}
  \definecolor{perf30}{HTML}{eee0e3}
  \definecolor{perf40}{HTML}{e8d5da}
  \definecolor{perf50}{HTML}{e2cbd1}
  \definecolor{perf60}{HTML}{dcc0c7}
  \definecolor{perf70}{HTML}{d5b3bb}
  \definecolor{perf80}{HTML}{cea5af}

  \definecolor{firstcolor}{HTML}{E8C8D0}
  \definecolor{secondcolor}{HTML}{E8D0D6}
  \definecolor{thirdcolor}{HTML}{E8DCDF}

  \definecolor{PaletteDeep}{HTML}{270A11}
  \definecolor{PaletteIndigo}{HTML}{561A29}
  \definecolor{PaletteMid}{HTML}{A72B4A}
  \definecolor{PaletteLight}{HTML}{EAA8B9}
  \definecolor{PaletteSilver}{HTML}{EFD4DB}
}{%
\ifthenelse{\equal{\themename}{blue}}{
  \newcommand{\thelogo}{}
  \definecolor{accent_primary}{HTML}{3C4A57}   %
  \definecolor{accent_cite}{HTML}{3E464E}      %
  \definecolor{bg_title}{HTML}{F4F5F6}         %
  \definecolor{bg_sepia}{HTML}{F5F6F7}         %
  \definecolor{bordercolor}{HTML}{CBD1D6}      %

  \definecolor{perf0}{HTML}{ffffff}
  \definecolor{perf10}{HTML}{f6f8fa}
  \definecolor{perf20}{HTML}{edf0f3}
  \definecolor{perf30}{HTML}{e2e7ec}
  \definecolor{perf40}{HTML}{d8dfe5}
  \definecolor{perf50}{HTML}{cfd6de}
  \definecolor{perf60}{HTML}{c5ced7}
  \definecolor{perf70}{HTML}{b9c4cf}
  \definecolor{perf80}{HTML}{acbac7}

  \definecolor{firstcolor}{HTML}{CED8E2}
  \definecolor{secondcolor}{HTML}{D4DCE4}
  \definecolor{thirdcolor}{HTML}{DEE2E6}

  \definecolor{PaletteDeep}{HTML}{0F1822}
  \definecolor{PaletteIndigo}{HTML}{25384C}
  \definecolor{PaletteMid}{HTML}{416991}
  \definecolor{PaletteLight}{HTML}{B4C9DE}
  \definecolor{PaletteSilver}{HTML}{D9E2EA}
}{%
\ifthenelse{\equal{\themename}{gold}}{
  \newcommand{\thelogo}{}
  \definecolor{accent_primary}{HTML}{564D3D}   %
  \definecolor{accent_cite}{HTML}{4D483F}      %
  \definecolor{bg_title}{HTML}{F6F5F4}         %
  \definecolor{bg_sepia}{HTML}{F7F6F5}         %
  \definecolor{bordercolor}{HTML}{D6D2CB}      %

  \definecolor{perf0}{HTML}{ffffff}
  \definecolor{perf10}{HTML}{faf8f6}
  \definecolor{perf20}{HTML}{f3f1ed}
  \definecolor{perf30}{HTML}{ebe8e3}
  \definecolor{perf40}{HTML}{e4e0d9}
  \definecolor{perf50}{HTML}{ddd8d0}
  \definecolor{perf60}{HTML}{d6d0c6}
  \definecolor{perf70}{HTML}{cec7ba}
  \definecolor{perf80}{HTML}{c6bdad}

  \definecolor{firstcolor}{HTML}{E2DBCE}
  \definecolor{secondcolor}{HTML}{E3DED5}
  \definecolor{thirdcolor}{HTML}{E6E3DE}

  \definecolor{PaletteDeep}{HTML}{211B10}
  \definecolor{PaletteIndigo}{HTML}{4A3D26}
  \definecolor{PaletteMid}{HTML}{8E7344}
  \definecolor{PaletteLight}{HTML}{DDCEB5}
  \definecolor{PaletteSilver}{HTML}{EAE4D9}
}{%
\ifthenelse{\equal{\themename}{green}}{
  \newcommand{\thelogo}{}
  \definecolor{accent_primary}{HTML}{2A4438}   %
  \definecolor{accent_cite}{HTML}{2A4438}      %
  \definecolor{bg_title}{HTML}{F4F6F5}         %
  \definecolor{bg_sepia}{HTML}{F5F7F6}         %
  \definecolor{bordercolor}{HTML}{C8D9D2}      %

  \definecolor{perf0}{HTML}{ffffff}
  \definecolor{perf10}{HTML}{f5faf8}
  \definecolor{perf20}{HTML}{ebf4f0}
  \definecolor{perf30}{HTML}{e0ede7}
  \definecolor{perf40}{HTML}{d5e6de}
  \definecolor{perf50}{HTML}{cbe0d6}
  \definecolor{perf60}{HTML}{c0d9cd}
  \definecolor{perf70}{HTML}{b3d1c3}
  \definecolor{perf80}{HTML}{a5c9b9}

  \definecolor{firstcolor}{HTML}{C8E8DB}
  \definecolor{secondcolor}{HTML}{D0E8DE}
  \definecolor{thirdcolor}{HTML}{DCE8E3}

  \definecolor{PaletteDeep}{HTML}{081810}
  \definecolor{PaletteIndigo}{HTML}{143026}
  \definecolor{PaletteMid}{HTML}{24604A}
  \definecolor{PaletteLight}{HTML}{A8EACF}
  \definecolor{PaletteSilver}{HTML}{D4EFE4}
}{
  \newcommand{\thelogo}{}
  \definecolor{accent_primary}{HTML}{3E2548}   %
  \definecolor{accent_cite}{HTML}{3E2548}      %
  \definecolor{bg_title}{HTML}{F6F4F6}         %
  \definecolor{bg_sepia}{HTML}{F7F5F7}         %
  \definecolor{bordercolor}{HTML}{D6C8D9}      %

  \definecolor{perf0}{HTML}{ffffff}
  \definecolor{perf10}{HTML}{f9f5fb}
  \definecolor{perf20}{HTML}{f2ebf5}
  \definecolor{perf30}{HTML}{ebe0ee}
  \definecolor{perf40}{HTML}{e4d5e8}
  \definecolor{perf50}{HTML}{ddcbe2}
  \definecolor{perf60}{HTML}{d6c0dc}
  \definecolor{perf70}{HTML}{ceb3d5}
  \definecolor{perf80}{HTML}{c6a5ce}

  \definecolor{firstcolor}{HTML}{E3C8E8}
  \definecolor{secondcolor}{HTML}{E4D0E8}
  \definecolor{thirdcolor}{HTML}{E6DCE8}

  \definecolor{PaletteDeep}{HTML}{150820}
  \definecolor{PaletteIndigo}{HTML}{2A1238}
  \definecolor{PaletteMid}{HTML}{5C2870}
  \definecolor{PaletteLight}{HTML}{DFA8EA}
  \definecolor{PaletteSilver}{HTML}{EBD4EF}
}}}}

\definecolor{LightCyan}{rgb}{.9, .95, 1.}
\definecolor{SDEblue}{RGB}{28, 58, 88}
\definecolor{cc1}{rgb}{1.0, 0.44, 0.37}
\definecolor{cc2}{rgb}{0.0, 0.2, 0.6}
\definecolor{cc3}{RGB}{255, 191, 0}
\definecolor{cc4}{RGB}{0, 128, 128}

\colorlet{TinaCrimson}{accent_primary}
\colorlet{CalGoldHex}{bg_title}
\colorlet{YaleBlue}{accent_cite}

\tcbset{titlebox/.style={colback=bg_title, colframe=bg_title,
    boxrule=0mm, arc=2mm, auto outer arc,
    left=5mm, right=5mm, top=5mm, bottom=5mm, enhanced}}

\newtcolorbox{abox}{colback=bg_sepia,colframe=accent_primary!60,
    boxrule=0.3mm,arc=2mm,left=5mm,right=5mm,top=3mm,bottom=3mm}

\hypersetup{
    colorlinks=true,
    urlcolor=accent_primary,
    citecolor=accent_cite,
    linkcolor=accent_primary
}

\usepackage{graphicx}
\usepackage{subcaption}
\usepackage{booktabs}
\usepackage{multirow}
\usepackage{colortbl}
\usepackage{amsmath}
\usepackage{amssymb}

\IfFileExists{generated/results_macros.tex}{
\providecommand{\ResultMainBaseQwenAgentVstar}{62.83\%}
\providecommand{\ResultMainBaseQwenAgentVstarValue}{62.83}

\providecommand{\ResultMainBaseQwenAgentHrBenchFourKValue}{54.75}

\providecommand{\ResultMainBaseQwenAgentHrBenchEightKValue}{50.00}

\providecommand{\ResultMainBaseQwenAgentMmStarValue}{58.33}

\providecommand{\ResultMainBaseQwenAgentCvBenchTwoDValue}{70.51}

\providecommand{\ResultMainBaseQwenAgentCvBenchThreeDValue}{68.67}

\providecommand{\ResultMainGThreeTwiSftNineKVstar}{71.73\%}
\providecommand{\ResultMainGThreeTwiSftNineKVstarValue}{71.73}

\providecommand{\ResultMainGFourBypassSftNineKVstar}{76.96\%}
\providecommand{\ResultMainGFourBypassSftNineKVstarValue}{76.96}

\providecommand{\ResultMainGThreeTwiSftSixtyFiveKVstarValue}{78.53}

\providecommand{\ResultMainGFourBypassSftSixtyFiveKVstarValue}{78.01}

\providecommand{\ResultMainGThreeTwiSftNineKHrBenchFourKValue}{63.00}

\providecommand{\ResultMainGThreeTwiSftNineKHrBenchEightKValue}{54.00}

\providecommand{\ResultMainGThreeTwiSftNineKMmStarValue}{56.20}

\providecommand{\ResultMainGThreeTwiSftNineKCvBenchTwoDValue}{74.48}

\providecommand{\ResultMainGThreeTwiSftNineKCvBenchThreeDValue}{66.92}

\providecommand{\ResultMainGFourBypassSftNineKMmStarValue}{56.93}

\providecommand{\ResultMainGFourBypassSftNineKHrBenchFourKValue}{71.50}

\providecommand{\ResultMainGFourBypassSftNineKHrBenchEightKValue}{67.00}

\providecommand{\ResultMainGFourBypassSftNineKCvBenchTwoDValue}{70.93}

\providecommand{\ResultMainGFourBypassSftNineKCvBenchThreeDValue}{68.42}

\providecommand{\ResultMainGThreeTwiSftSixtyFiveKHrBenchFourKValue}{69.25}

\providecommand{\ResultMainGThreeTwiSftSixtyFiveKHrBenchEightKValue}{64.38}

\providecommand{\ResultMainGThreeTwiSftSixtyFiveKMmStarValue}{61.33}

\providecommand{\ResultMainGThreeTwiSftSixtyFiveKCvBenchTwoDValue}{73.37}

\providecommand{\ResultMainGThreeTwiSftSixtyFiveKCvBenchThreeDValue}{72.17}

\providecommand{\ResultMainGFourBypassSftSixtyFiveKHrBenchFourKValue}{73.12}

\providecommand{\ResultMainGFourBypassSftSixtyFiveKHrBenchEightKValue}{67.50}

\providecommand{\ResultMainGFourBypassSftSixtyFiveKMmStarValue}{62.73}

\providecommand{\ResultMainGFourBypassSftSixtyFiveKCvBenchTwoDValue}{72.32}

\providecommand{\ResultMainGFourBypassSftSixtyFiveKCvBenchThreeDValue}{73.67}

\providecommand{\ResultMainBaseQwenDirectMcVstarValue}{80.63}

\providecommand{\ResultMainBaseQwenDirectMcHrBenchFourKValue}{71.25}

\providecommand{\ResultMainBaseQwenDirectMcHrBenchEightKValue}{68.62}

\providecommand{\ResultMainBaseQwenDirectMcMmeRealWorldValue}{61.42}

\providecommand{\ResultMainBaseQwenDirectMcMmStarValue}{62.80}

\providecommand{\ResultMainBaseQwenDirectMcCvBenchTwoDValue}{75.45}

\providecommand{\ResultMainBaseQwenDirectMcCvBenchThreeDValue}{84.83}

\providecommand{\ResultMainBaseQwenDirectMcBlinkValue}{54.45}

\providecommand{\ResultMainBaseQwenAgentBlinkValue}{49.29}

\providecommand{\ResultMainBaseQwenDirectMcChartqaValue}{86.36}

\providecommand{\ResultMainBaseQwenAgentChartqaValue}{47.60}

\providecommand{\ResultMainBaseQwenDirectMcPixmocountValue}{64.75}

\providecommand{\ResultMainBaseQwenAgentPixmocountValue}{64.09}

\providecommand{\ResultMainGThreeTwiSftSixtyFiveKChartqaValue}{64.48}

\providecommand{\ResultMainGThreeTwiSftSixtyFiveKPixmocountValue}{56.93}

\providecommand{\ResultMainGThreeTwiSftSixtyFiveKBlinkValue}{53.13}

\providecommand{\ResultMainGFourBypassSftSixtyFiveKChartqaValue}{66.24}

\providecommand{\ResultMainGFourBypassSftSixtyFiveKPixmocountValue}{55.04}

\providecommand{\ResultMainGThreeTwiSftSixtyFiveKMmerwValue}{61.01}

\providecommand{\ResultMainGFourBypassSftSixtyFiveKMmerwValue}{62.05}

\providecommand{\ResultMainGFourBypassSftSixtyFiveKBlinkValue}{54.29}

\providecommand{\ResultMainGThreeTwiSftNineKBlinkValue}{49.50}

\providecommand{\ResultMainGFourBypassSftNineKBlinkValue}{51.29}

\providecommand{\ResultMainGFourBypassSftNineKMmerwValue}{61.02}

\providecommand{\ResultMainGFourBypassSftNineKChartqaValue}{53.84}

\providecommand{\ResultMainGFourBypassSftNineKPixmocountValue}{50.05}

\providecommand{\ResultMainGThreeTwiSftNineKChartqaValue}{48.36}

\providecommand{\ResultMainGThreeTwiSftNineKPixmocountValue}{53.53}

\providecommand{\ResultMainGThreeTwiSftNineKMmerwValue}{57.73}

\providecommand{\ResultMainBaseQwenAgentMmerwValue}{58.04}

\providecommand{\ResultMainGFourBypassRlSixtyFiveKVstarValue}{81.68}

\providecommand{\ResultMainGFourBypassRlSixtyFiveKHrBenchFourKValue}{73.88}

\providecommand{\ResultMainGFourBypassRlSixtyFiveKHrBenchEightKValue}{70.62}

\providecommand{\ResultMainGFourBypassRlSixtyFiveKMmStarValue}{65.53}

\providecommand{\ResultMainGFourBypassRlSixtyFiveKCvBenchTwoDValue}{67.32}

\providecommand{\ResultMainGFourBypassRlSixtyFiveKCvBenchThreeDValue}{56.83}

\providecommand{\ResultMainGFourBypassRlSixtyFiveKBlinkValue}{45.98}

\providecommand{\ResultMainGFourBypassRlSixtyFiveKChartQAValue}{85.20}

\providecommand{\ResultMainGFourBypassRlSixtyFiveKMmerwValue}{64.51}

\providecommand{\ResultMainGFourBypassRlSixtyFiveKPixmoCountValue}{57.12}

\providecommand{\ResultMainDevTwoSftVstarValue}{74.87}

\providecommand{\ResultMainDevTwoSftHrbenchFourKValue}{69.88}

\providecommand{\ResultMainBaseDirectmcCharxivReasValue}{22.00}

\providecommand{\ResultMainBaseAgentCharxivReasValue}{20.30}

\providecommand{\ResultMainGThreeTwiSftSixtyFiveKCharxivReasValue}{39.80}

\providecommand{\ResultMainGFourBypassSftSixtyFiveKCharxivReasValue}{41.90}

\providecommand{\ResultMainGFourBypassRlSixtyFiveKCharxivReasValue}{29.30}

\providecommand{\ResultMainDevTwoSftHrbenchEightKValue}{62.38}

\providecommand{\ResultMainDevTwoSftMmstarValue}{62.80}

\providecommand{\ResultMainDevTwoSftCvbenchTwoDValue}{74.48}

\providecommand{\ResultMainDevTwoSftChartqaValue}{80.20}

\providecommand{\ResultMainDevTwoSftPixmocountValue}{54.19}

\providecommand{\ResultMainDevTwoSftCvbenchThreeDValue}{78.00}

\providecommand{\ResultMainDevTwoSftBlinkValue}{50.03}

\providecommand{\ResultMainDevTwoSftCharxivReasValue}{36.30}

\providecommand{\ResultMainDevTwoRlVstarValue}{83.25}

\providecommand{\ResultMainDevTwoRlHrbenchFourKValue}{76.00}

\providecommand{\ResultMainDevTwoRlCharxivReasValue}{39.10}

\providecommand{\ResultMainDevTwoRlHrbenchEightKValue}{70.75}

\providecommand{\ResultMainDevTwoRlMmstarValue}{63.00}

\providecommand{\ResultMainDevTwoRlCvBenchTwoDValue}{77.05}

\providecommand{\ResultMainDevTwoRlCvBenchThreeDValue}{80.25}

\providecommand{\ResultMainDevTwoRlPixmocountValue}{57.40}

\providecommand{\ResultMainDevTwoRlBlinkValue}{54.02}

\providecommand{\ResultMainGThreeTwiSftNineKCharxivReasValue}{26.40}

\providecommand{\ResultMainGFourBypassSftNineKCharxivReasValue}{26.20}

\providecommand{\ResultSecFourDegradationLOneVstarValue}{64.92}

\providecommand{\ResultSecFourDegradationLOneHrBenchFourKValue}{65.50}

\providecommand{\ResultSecFourDegradationLOneCvBenchTwoDValue}{55.08}

\providecommand{\ResultSecFourDegradationLTwoVstarValue}{67.02}

\providecommand{\ResultSecFourDegradationLTwoHrBenchFourKValue}{72.12}

\providecommand{\ResultSecFourDegradationLTwoCvBenchTwoDValue}{59.81}

\providecommand{\ResultAppDegradationLThreeVstarValue}{79.06}

\providecommand{\ResultAppDegradationLThreeHrBenchFourKValue}{72.62}

\providecommand{\ResultAppDegradationLThreeCvBenchTwoDValue}{63.21}

\providecommand{\ResultMainDevTwoSftMmerwValue}{60.41}

\providecommand{\ResultMainDevTwoRlMmerwValue}{64.35}

\providecommand{\ResultMainDevTwoRlChartqaValue}{86.08}

\providecommand{\ResultMainGThreeOverBaseSftSixtyFiveKVstar}{+15.70\,pp}

\providecommand{\ResultMainGapSftNineKVstar}{+5.23\,pp}

\providecommand{\ResultMainGapSftSixtyFiveKVstar}{-0.52\,pp}

}{}

\newcommand{\methodnameplain}{TextCall}
\newcommand{\methodname}{\textbf{\methodnameplain}}
\newcommand{\methodnamebold}{\methodname}
\newcommand{\methodnamefull}{\methodname{} (call-but-no-return)}
\newcommand{\weak}[1]{#1}

\newcommand{\jigsawfirstpagewidth}{0.75\textwidth}
\newcommand{\jigsawsecondpagewidth}{0.785\textwidth}

\newtcolorbox{summarybox}[1][]{
  enhanced jigsaw,
  colback=bg_sepia,
  colframe=accent_primary!60,
  arc=5pt,
  boxsep=5pt,
  left=10pt,
  right=10pt,
  top=4pt,
  bottom=4pt,
  boxrule=0.8pt,
  fonttitle=\bfseries,
  #1
}

\title{Thinking With Tools, Not With Pixels: \\ Tool Calls as Text Scaffolds for Visual Reasoning}
\runningtitle{Thinking With Tools, Not With Pixels}
\author{
  Jiahao Shao\textsuperscript{1},
  Yuanbo Yang\textsuperscript{4},
  Yiyi Liao\textsuperscript{4},
  Yujun Shen\textsuperscript{3},
  Ceyuan Yang\textsuperscript{2,$\dagger$},
  Yinghao Xu\textsuperscript{1,$\dagger$}\\
  \normalfont\small
  \textsuperscript{1}\,Hong Kong University of Science and Technology \quad
  \textsuperscript{2}\,The Chinese University of Hong Kong\\\vspace{-0.45em}
  \textsuperscript{3}\,Ant Group \quad
  \textsuperscript{4}\,Zhejiang University\\\vspace{-0.15em}
  \href{https://textcall.github.io/}{\texttt{Project Page: textcall.github.io}}
}

\begin{document}

\begin{abstract}
Tool-augmented vision-language models increasingly ``think with images'': they call crop, zoom, or code tools and reason over the returned pixels.
However, recent work using blind tests, gain decompositions, and attention analyses has shown that returned images contribute little, raising the question: if pixels do not carry the gain, what does?
We hypothesize that the load-bearing signal is the structured text emitted before any returned pixel arrives: tool name, coordinates, target description, and intent.
This textual scaffold encodes where to look and what to find.
We introduce \methodnamefull{} to test this: it keeps the scaffold but replaces returned images with the text placeholder \texttt{[Image output skipped]}.
Three studies support the hypothesis.
(i)~\emph{Non-necessity of returned pixels}: across LoRA, full fine-tuning, and RL, \methodname{} matches or exceeds full thinking-with-images; under RL it preserves tool use at the reported checkpoint, avoiding the failure mode where, under matched settings, seeing the returned image causes the model to stop calling tools and answer directly.
(ii)~\emph{Sufficiency of the scaffold}: on matched training queries, scaffold-only input yields equivalent accuracy to returned-image input.
(iii)~\emph{Component specificity}: decomposing the scaffold into reasoning text and spatial code shows both components contribute, with the dominant one varying by task.
Together these results support the \textbf{Tool-Call Scaffold Hypothesis}: in current thinking-with-images distributions, the active signal is the structured text emitted at tool-call time; the returned image is a redundant carrier.
\methodname{} preserves accuracy while reducing latency by 29--46\% and eliminating tool-execution API calls.
Our claims hold for current thinking-with-images benchmarks; constructing tasks where pixels are genuinely load-bearing remains an open direction.

\end{abstract}

\maketitle
\begingroup
\renewcommand{\thefootnote}{\fnsymbol{footnote}}
\footnotetext[2]{Equal advising.}
\endgroup
\suppressfloats[t]

\section{Introduction}
\label{sec:intro}

\begin{figure}[t]
\centering
\includegraphics[width=\textwidth]{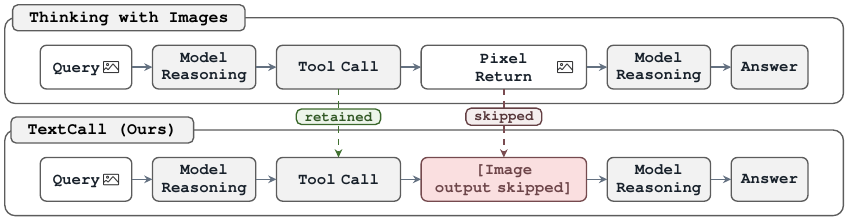}
\caption{\textbf{\methodname{} preserves the tool-call scaffold and removes only the returned pixels.} Both pipelines emit a structured textual scaffold at tool-call time. Thinking-with-images then consumes the returned pixels, while \methodname{} receives the text placeholder \texttt{[Image output skipped]}. The retained scaffold channel and skipped pixel channel define the \textbf{controlled carrier swap} used throughout the paper.}
\label{fig:teaser}
\end{figure}

Tool-augmented vision-language models (VLMs) increasingly think with images: they call crop, zoom, or code tools and reason over the returned pixels~\citep{deepeyes,pixelreasoner,deepeyesv2,vigorl,zwz,visualsketchpad}.
This ``thinking-with-images'' (TWI) paradigm---a form of multimodal reasoning that interleaves language generation with visual tool use---has driven steady gains on perception-heavy benchmarks and is now the engineering default at industrial scale~\citep{openaitwi,qwen3vl,glm5vturbo}.
Yet a growing set of diagnostics challenges the premise that returned images carry the reasoning signal.
At the most direct level, replacing returned images with noise drops V*Bench accuracy by only $0.52$ percentage points (pp)~\citep{faithfulness}, removing images at inference even raises MathVista accuracy by $3.5$\,pp~\citep{deltas}, and only $57\%$ of returned crops actually contain the target object~\citep{codev}.
More structurally, over $70\%$ of post-RL gains arise from intrinsic capability rather than tool use~\citep{med}, and text-only self-calling chains of thought already outperform their pixel-interleaved counterparts~\citep{scot}.
Taken together, these diagnostics leave the positive mechanism unresolved: if returned pixels are not carrying the gain, what signal is?

We hypothesize that the missing causal signal is the structured text the model emits \emph{before} any returned pixel arrives: the tool name, target coordinates, target description, and call intent (e.g., ``crop the lower-left region to inspect the handbag'') collectively form a \emph{textual scaffold} that already encodes where to look and what to find.
If this hypothesis holds, replacing the returned pixels with a non-visual carrier should preserve performance.
We call this intervention \methodnamefull{}: keep the tool-call scaffold, but replace the returned image with the text placeholder \texttt{[Image output skipped]}.

We test this hypothesis with three studies.
1) \textbf{Non-necessity of returned pixels.} Across small-scale LoRA and large-scale full fine-tuning, \methodname{} matches or exceeds full thinking-with-images training on a six-benchmark core suite; after Reinforcement Learning, \methodname{} maintains active tool use at the reported checkpoint while the matched thinking-with-images run collapses to direct answering.
2) \textbf{Sufficiency of the scaffold.} On ${\sim}1{,}000$ matched training queries, the scaffold-only input yields equivalent task accuracy to the returned-image input.
3) \textbf{Component specificity.} The scaffold is not arbitrary text: decomposing it into reasoning text (intent and target description) and spatial code (crop coordinates) shows that both components matter, though coordinate grounding matters more on some benchmarks than others.

These three angles converge on the \textbf{Tool-Call Scaffold Hypothesis}: in current thinking-with-images distributions, the load-bearing signal is the structured text emitted at tool-call time, while the returned image is often only the carrier being replaced.
In practice, \methodname{} preserves accuracy while reducing per-sample latency by 29--46\%, eliminating tool-execution overhead, and removing multi-turn image-token injection from the architecture.

Our contributions are as follows:
\begin{itemize}
    \item We introduce a carrier-swap ablation that isolates returned pixels from the tool-call scaffold and tests pixel necessity across LoRA, full fine-tuning, and RL settings.
    \item We diagnose why the swap works with a paired scaffold-vs-image audit and content-decomposition ablations that identify which scaffold components matter.
    \item We instantiate the principle as \methodnamebold{}, achieving parity-or-better accuracy with 29--46\% lower latency, zero tool-execution overhead, and no multi-turn image-token injection.
\end{itemize}

We scope our claims to current thinking-with-images training distributions and the evaluated benchmark suite; future tasks may make returned pixels genuinely load-bearing rather than redundant.
This boundary points to a concrete benchmark need: settings where the scaffold alone cannot substitute for the visual return.

\section{Related Work}
\label{sec:related}

\paragraph{Thinking-with-images: pixels as the unit of visual thought.}
Thinking-with-images (TWI) was introduced at industrial scale by OpenAI o3/o4-mini~\citep{openaitwi}, whose chain-of-thought natively crops, zooms, and rotates input images via tool calls. Canonical crop-and-zoom TWI systems share an implicit assumption: the \emph{returned pixel} aids subsequent reasoning---the pixel return is the unit of visual thought. Earlier SFT-based manipulation-chain systems---CogCoM~\citep{cogcom} with intrinsic visual operations and Chain-of-Spot~\citep{chainofspot} with ROI-conditioned re-encoding---already operationalized this pixel-return assumption before the RL-driven wave. V*~\citep{vstar} introduces a visual search tool, DeepEyes~\citep{deepeyes} adds crop-based search with RL, and DeepEyes V2~\citep{deepeyesv2} scales the recipe with a cold-start corpus and multi-turn agentic RL~\citep{practitionerguide, ragen, prorl}; Pixel Reasoner~\citep{pixelreasoner} extends pixel-space operations with curiosity-driven RL; ViGoRL~\citep{vigorl} combines MCTS-grounded trajectories with GRPO. A wave of concurrent systems further expands the design space with interleaved vision-language reasoning, reward shaping for tool use, and domain-specific adaptations~\citep{openthinkimg, vtoolr1, simpleo3, minio3, verltool, vagen, geoeyes, pebr, thyme, skyworkr1v4, rewardshaping, cm2, tapo}. The paradigm has also reached industrial scale: Qwen3-VL~\citep{qwen3vl} and GLM-5V-Turbo~\citep{glm5vturbo} train native zoom/crop toolsets via cold-start SFT and tool-integrated RL. An adjacent line distills or bypasses the returned pixel altogether: Zooming without Zooming~\citep{zwz} collapses multi-turn zoom into a single forward pass, and Seg-Zero~\citep{segzero} routes text-only positional prompts to an external segmentation model whose mask outputs are not re-encoded into the reasoning stream---a concurrent RL instance of decoupling tool calls from pixel feedback; see~\citet{twisurvey} for a survey. Our work tests whether the pixel-return assumption holds by training without it.

\paragraph{Diagnoses of pixel necessity.}
Prior evidence challenges the pixel-as-unit reading from several angles (full table in Appendix~\ref{app:diagnostic_table}). Inference-time interventions show that returned pixels carry surprisingly little causal weight: replacing crops with noise drops accuracy by only $0.52$\,pp while corrupting text drops it by $13$\,pp~\citep{faithfulness}, removing images can even raise accuracy~\citep{deltas}, over $70\%$ of post-RL gains arise from intrinsic reasoning rather than tool use~\citep{med, pebr}, and only $57\%$ of crops contain the target object~\citep{codev}. Training- and data-side evidence corroborates: original-image attention rewards recover most gains without tool-output signal~\citep{sayo}, zoom benefits can be distilled into single-pass inference~\citep{zwz}, lighter spatial mechanisms such as implicit re-focus tokens~\citep{lookback} and bbox-coordinate reasoning chains~\citep{grit} achieve grounding without pixel returns, zero-vision SFT activates visual reasoning without visual data~\citep{kimik25}, and text-only self-calling CoT outperforms pixel-interleaved iMCoT~\citep{scot}. Meanwhile, \emph{divergent} repair strategies---aligning visual actions~\citep{mapo}, reducing blind invocations~\citep{actwisely, geoeyes}, step-level verification~\citep{processrewardtwi}---improve performance through orthogonal levers, further suggesting that the role of returned images remains poorly isolated. What is missing is a \emph{matched training-time carrier swap}: models that never see image feedback during training, removing the train--test mismatch confound that clouds inference-time diagnostics. \methodname{} provides this counterpart and isolates the role of the textual scaffold.

\paragraph{The faithfulness gap in language-space reasoning.}
A parallel line of work asks whether LLMs and VLMs truly reason over their decoded traces. Chain-of-thought faithfulness studies show that reasoning traces frequently diverge from the model's internal decision process~\citep{turpin2023, lanham2023, anthropicfaithful, lietome}; in VLMs specifically, image-based biases influence answers yet rarely surface in CoT~\citep{vlmsblind, breakingchain, facte, acloserlook}, RL training improves accuracy while \emph{degrading} CoT faithfulness~\citep{cotrobustness}, and thinking-mode reasoning amplifies text bias rather than strengthening visual grounding~\citep{vfat}. Visual tokens become redundant after early layers~\citep{fastv, llavamini}, and latent-space alternatives show reasoning need not even be decoded~\citep{coconut, plat, thinkingstates}. From the training side, textual slow-thinking data transfers to multimodal reasoning~\citep{virgo}, though extended chains risk ``visual anchor drifting''~\citep{whenthinkighurts, journeybeforedest, revisitlongcot}. Unlike latent-reasoning work, we study the \emph{decoded} tool-call scaffold emitted before the visual observation: our claim is that this observable scaffold is the load-bearing signal under a controlled carrier swap.

\paragraph{Structured text as the reasoning medium.}
Structured intermediate text---chain-of-thought~\citep{cot, scratchpad}, self-consistency extensions~\citep{selfconsistency, leasttomost, tot}, reasoning-trace RL~\citep{deepseekr1}, and multimodal CoT~\citep{multimodalcot, visualcot, llavacot}---has been the locus of capability gains in both LLMs and VLMs. MM-CoT~\citep{mmcotbench} benchmarks whether visual CoT is grounded in images; Thinking in Space~\citep{thinkinginspace} identifies the spatial reasoning boundary where textual CoT fails. Our work extends this thread to tool-augmented VLMs, providing the training-time ablation that prior CoT studies lack.

\paragraph{Tool calls as actions vs.\ as thought primitives.}
Tool-augmented LLMs traditionally treat the tool call as an \emph{action} whose returned observation drives the next step~\citep{react, toolformer, vipergpt, visprog, chameleon, hugginggpt}. Adjacent visual-prompting work shows that Set-of-Mark's symbolic mark identifiers enable GPT-4V to reference spatial regions in its textual output~\citep{setofmark}. VisionThink~\citep{visionthink} finds that 50\% of pixel tokens can be skipped without loss---evidence for selective pixel redundancy. Concurrent work also explores alternative carriers between text and re-encoded pixels: v1~\citep{v1pointcopy} learns to copy visual token embeddings back into the reasoning stream rather than re-encoding cropped images. These results, together with counterfactual studies probing whether reasoning depends on its ostensible substrate~\citep{reasoningwithoutthinking, reasoningorreciting, gsmsymbolic}, suggest a re-interpretation of the tool call as a \emph{thought primitive}---a structured token sequence whose value may be largely independent of the observation it triggers. Our pixel-return ablations provide direct training-time evidence for this re-interpretation within the TWI paradigm.

\begin{figure}[!t]
    \centering
    \includegraphics[width=\textwidth]{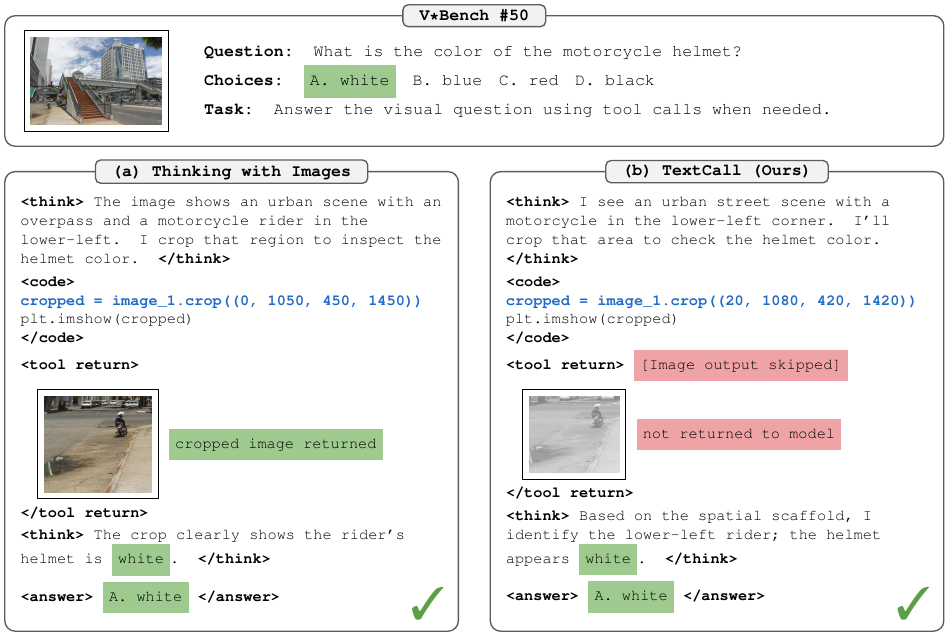}
    \caption{\textbf{\methodname{} preserves the tool-call scaffold while removing the image return.} On V*Bench \#50, both thinking-with-images and \methodname{} identify the white motorcycle helmet. Thinking-with-images receives the cropped image after issuing the crop call, whereas \methodname{} emits a closely matched crop call but receives only \texttt{[Image output skipped]}. The example illustrates the controlled carrier swap used in Section~\ref{sec:causal}: the tool-call scaffold is preserved, while the post-call pixel carrier is removed. Additional examples in Appendix~\ref{sec:scaffold-gallery}.}
    \label{fig:scaffold-overview}
\end{figure}

\section{\methodname{} Matches or Exceeds Thinking-with-Images}
\label{sec:causal}

Tool-call training improves VLMs by teaching them to emit structured calls that localize and query visual evidence. We ask whether this gain requires the returned pixels, or whether the pre-return tool-call scaffold is sufficient. Prior diagnostics mostly remove or corrupt tool outputs only at inference time, leaving a train--test mismatch: the model was trained to expect images but evaluated without them. \methodname{} answers the training-time version of this question: it preserves the tool call and execution confirmation, but replaces every returned image with the fixed sentinel ``\texttt{[Image output skipped]}'' during both training and inference. Across SFT scales, \methodname{} matches or exceeds thinking-with-images on the core suite. Under GRPO, \methodname{} maintains active tool use at the reported checkpoint; the matched thinking-with-images RL run collapses to direct answering and is analyzed separately as training-dynamics evidence.

\begin{table}[t]
\centering
\scriptsize
\setlength{\tabcolsep}{3pt}
\caption{\textbf{\methodname{} matches or exceeds thinking-with-images across training scales.} Symbol annotations: $^\dagger$ indicates results reported in the original papers; $^\ddagger$ indicates our re-evaluation of official checkpoints; $^\diamond$ marks a data-size estimate from Figure~9 in DeepEyes~V2~\citep{deepeyesv2}; $^\text{L}$ marks LoRA fine-tuning; $^{*}$ marks GRPO fine-tuning; --- indicates unreported results.}
\label{tab:main_results}
\resizebox{\textwidth}{!}{%
\begin{tabular}{@{}llcccccccccccc@{}}
\toprule
& & & \multicolumn{4}{c}{High-Res Perception} & \multicolumn{3}{c}{Spatial \& Counting} & \multicolumn{2}{c}{General Vision} & \multicolumn{2}{c}{Charts \& Figures} \\
\cmidrule(lr){4-7}\cmidrule(lr){8-10}\cmidrule(lr){11-12}\cmidrule(l){13-14}
& Data & Eval & V$^{\!*}$ & HR-4K & HR-8K & MME-RW & CV-2D & CV-3D & PixCnt & MMStar & BLINK & ChartQA & CharXiv \\
\midrule
\multicolumn{14}{l}{\textit{Tool-using models (Agent Mode)}$^\dagger$} \\
DeepEyes V2~\citep{deepeyesv2} & ${\sim}$358K$^\diamond$ & Agent & 81.8 & 77.9 & 73.8 & 64.9 & --- & --- & --- & --- & --- & --- & --- \\
Pixel-Reasoner~\citep{pixelreasoner} & --- & Agent & 84.3 & 74.0 & 66.9 & 64.4 & --- & --- & --- & --- & --- & --- & --- \\
\midrule
\multicolumn{14}{l}{\textit{Tool-using models (ours re-eval on official ckpt, same pipeline)}$^\ddagger$} \\
DeepEyes V2 SFT~\citep{deepeyesv2} & ${\sim}$358K$^\diamond$ & Agent & \ResultMainDevTwoSftVstarValue{} & \ResultMainDevTwoSftHrbenchFourKValue{} & \ResultMainDevTwoSftHrbenchEightKValue{} & \ResultMainDevTwoSftMmerwValue{} & \ResultMainDevTwoSftCvbenchTwoDValue{} & \ResultMainDevTwoSftCvbenchThreeDValue{} & \ResultMainDevTwoSftPixmocountValue{} & \ResultMainDevTwoSftMmstarValue{} & \ResultMainDevTwoSftBlinkValue{} & \ResultMainDevTwoSftChartqaValue{} & \ResultMainDevTwoSftCharxivReasValue{} \\
DeepEyes V2 RL~\citep{deepeyesv2} & ${\sim}$83K$^\diamond$ & Agent & \ResultMainDevTwoRlVstarValue{} & \ResultMainDevTwoRlHrbenchFourKValue{} & \ResultMainDevTwoRlHrbenchEightKValue{} & \ResultMainDevTwoRlMmerwValue{} & \ResultMainDevTwoRlCvBenchTwoDValue{} & \ResultMainDevTwoRlCvBenchThreeDValue{} & \ResultMainDevTwoRlPixmocountValue{} & \ResultMainDevTwoRlMmstarValue{} & \ResultMainDevTwoRlBlinkValue{} & \ResultMainDevTwoRlChartqaValue{} & \ResultMainDevTwoRlCharxivReasValue{} \\
\midrule
\multicolumn{14}{l}{\textit{Ours}} \\
\quad Base (Qwen2.5-VL-7B) & --- & Direct & \ResultMainBaseQwenDirectMcVstarValue{} & \ResultMainBaseQwenDirectMcHrBenchFourKValue{} & \ResultMainBaseQwenDirectMcHrBenchEightKValue{} & \ResultMainBaseQwenDirectMcMmeRealWorldValue{} & \textbf{\ResultMainBaseQwenDirectMcCvBenchTwoDValue{}} & \textbf{\ResultMainBaseQwenDirectMcCvBenchThreeDValue{}} & \textbf{\ResultMainBaseQwenDirectMcPixmocountValue{}} & \ResultMainBaseQwenDirectMcMmStarValue{} & \textbf{\ResultMainBaseQwenDirectMcBlinkValue{}} & \textbf{\ResultMainBaseQwenDirectMcChartqaValue{}} & \ResultMainBaseDirectmcCharxivReasValue{} \\
\quad Base (Qwen2.5-VL-7B) & --- & Agent & \weak{\ResultMainBaseQwenAgentVstarValue{}} & \weak{\ResultMainBaseQwenAgentHrBenchFourKValue{}} & \weak{\ResultMainBaseQwenAgentHrBenchEightKValue{}} & \weak{\ResultMainBaseQwenAgentMmerwValue{}} & \weak{\ResultMainBaseQwenAgentCvBenchTwoDValue{}} & \weak{\ResultMainBaseQwenAgentCvBenchThreeDValue{}} & \ResultMainBaseQwenAgentPixmocountValue{} & \ResultMainBaseQwenAgentMmStarValue{} & \weak{\ResultMainBaseQwenAgentBlinkValue{}} & \weak{\ResultMainBaseQwenAgentChartqaValue{}} & \ResultMainBaseAgentCharxivReasValue{} \\
\quad thinking-with-images SFT & 9.5K$^\text{L}$ & Agent & \weak{\ResultMainGThreeTwiSftNineKVstarValue{}} & \weak{\ResultMainGThreeTwiSftNineKHrBenchFourKValue{}} & \weak{\ResultMainGThreeTwiSftNineKHrBenchEightKValue{}} & \weak{\ResultMainGThreeTwiSftNineKMmerwValue{}} & \weak{\ResultMainGThreeTwiSftNineKCvBenchTwoDValue{}} & \weak{\ResultMainGThreeTwiSftNineKCvBenchThreeDValue{}} & \weak{\ResultMainGThreeTwiSftNineKPixmocountValue{}} & \weak{\ResultMainGThreeTwiSftNineKMmStarValue{}} & \weak{\ResultMainGThreeTwiSftNineKBlinkValue{}} & \weak{\ResultMainGThreeTwiSftNineKChartqaValue{}} & \ResultMainGThreeTwiSftNineKCharxivReasValue{} \\
\quad \methodnamebold{} SFT & 9.5K$^\text{L}$ & \methodnameplain{} & \weak{\ResultMainGFourBypassSftNineKVstarValue{}} & \ResultMainGFourBypassSftNineKHrBenchFourKValue{} & \ResultMainGFourBypassSftNineKHrBenchEightKValue{} & \ResultMainGFourBypassSftNineKMmerwValue{} & \weak{\ResultMainGFourBypassSftNineKCvBenchTwoDValue{}} & \weak{\ResultMainGFourBypassSftNineKCvBenchThreeDValue{}} & \weak{\ResultMainGFourBypassSftNineKPixmocountValue{}} & \weak{\ResultMainGFourBypassSftNineKMmStarValue{}} & \ResultMainGFourBypassSftNineKBlinkValue{} & \weak{\ResultMainGFourBypassSftNineKChartqaValue{}} & \ResultMainGFourBypassSftNineKCharxivReasValue{} \\
\quad thinking-with-images SFT & 65K & Agent & \ResultMainGThreeTwiSftSixtyFiveKVstarValue{} & \ResultMainGThreeTwiSftSixtyFiveKHrBenchFourKValue{} & \weak{\ResultMainGThreeTwiSftSixtyFiveKHrBenchEightKValue{}} & \ResultMainGThreeTwiSftSixtyFiveKMmerwValue{} & \ResultMainGThreeTwiSftSixtyFiveKCvBenchTwoDValue{} & \weak{\ResultMainGThreeTwiSftSixtyFiveKCvBenchThreeDValue{}} & \weak{\ResultMainGThreeTwiSftSixtyFiveKPixmocountValue{}} & \ResultMainGThreeTwiSftSixtyFiveKMmStarValue{} & \ResultMainGThreeTwiSftSixtyFiveKBlinkValue{} & \weak{\ResultMainGThreeTwiSftSixtyFiveKChartqaValue{}} & \ResultMainGThreeTwiSftSixtyFiveKCharxivReasValue{} \\
\quad \methodnamebold{} SFT & 65K & \methodnameplain{} & \ResultMainGFourBypassSftSixtyFiveKVstarValue{} & \ResultMainGFourBypassSftSixtyFiveKHrBenchFourKValue{} & \ResultMainGFourBypassSftSixtyFiveKHrBenchEightKValue{} & \ResultMainGFourBypassSftSixtyFiveKMmerwValue{} & \weak{\ResultMainGFourBypassSftSixtyFiveKCvBenchTwoDValue{}} & \weak{\ResultMainGFourBypassSftSixtyFiveKCvBenchThreeDValue{}} & \weak{\ResultMainGFourBypassSftSixtyFiveKPixmocountValue{}} & \ResultMainGFourBypassSftSixtyFiveKMmStarValue{} & \ResultMainGFourBypassSftSixtyFiveKBlinkValue{} & \weak{\ResultMainGFourBypassSftSixtyFiveKChartqaValue{}} & \textbf{\ResultMainGFourBypassSftSixtyFiveKCharxivReasValue{}} \\
\quad \methodnamebold{} SFT+RL$^{*}$ & 77K & \methodnameplain{} & \textbf{\ResultMainGFourBypassRlSixtyFiveKVstarValue{}} & \textbf{\ResultMainGFourBypassRlSixtyFiveKHrBenchFourKValue{}} & \textbf{\ResultMainGFourBypassRlSixtyFiveKHrBenchEightKValue{}} & \textbf{\ResultMainGFourBypassRlSixtyFiveKMmerwValue{}} & \ResultMainGFourBypassRlSixtyFiveKCvBenchTwoDValue{} & \ResultMainGFourBypassRlSixtyFiveKCvBenchThreeDValue{} & \ResultMainGFourBypassRlSixtyFiveKPixmoCountValue{} & \textbf{\ResultMainGFourBypassRlSixtyFiveKMmStarValue{}} & \ResultMainGFourBypassRlSixtyFiveKBlinkValue{} & \ResultMainGFourBypassRlSixtyFiveKChartQAValue{} & \ResultMainGFourBypassRlSixtyFiveKCharxivReasValue{} \\
\bottomrule
\end{tabular}%
}
\end{table}
\subsection{\methodname{} as a Training-Aligned Carrier Swap}
\label{sec:method}

In standard thinking-with-images training, the model receives two coupled signals: it learns to emit a structured tool call, and then it consumes the image returned by that call. \methodnamebold{} keeps the first signal intact and changes only the second. After the tool executes, the returned image is replaced with ``\texttt{[Image output skipped]}.'' The model still generates the full tool call instruction and observes the execution confirmation, but never receives the visual output. Critically, this is applied \emph{during training}---the model is never trained to expect images. See Figure~\ref{fig:teaser}.

This intervention is designed as a training-aligned carrier swap rather than a test-time ablation. It changes a single observable channel: the post-call image is replaced by a fixed text sentinel, while the model still produces the same tool-call scaffold and is evaluated under the protocol it was trained to follow. This avoids the main ambiguity of inference-time image removal~\citep{vstar, setofmark, deepeyes, faithfulness, deltas}, where a performance drop may reflect either genuine dependence on returned pixels or a train--inference distribution shift. Appendix~\ref{sec:method_return_ablation} details the return-text design and its scope.

\subsection{Experimental Setup}

\paragraph{Model.} We use Qwen2.5-VL-7B-Instruct~\citep{qwen25vl} (building on the dynamic-resolution architecture of Qwen2-VL~\citep{qwen2vl}) throughout. This is a state-of-the-art open VLM comparable to GPT-4V~\citep{gpt4} and Gemini~\citep{gemini} on standard benchmarks, while being amenable to full fine-tuning at 7B scale. We do not use Qwen3-VL as a backbone because its report already includes multimodal Long-CoT and tool-integrated thinking-with-images post-training~\citep{qwen3vl}, which would confound the carrier-swap intervention.

\paragraph{Training data.} SFT experiments use the DeepEyes V2 dataset (65K multi-turn tool-augmented trajectories; smaller subsets for ablations).

\paragraph{Benchmarks.} We evaluate on 11 perception/recognition benchmarks: a \emph{core} suite of six---\emph{V*Bench}~\citep{vstar} (191 samples), \emph{HR-Bench-4K} and \emph{HR-Bench-8K}~\citep{hrbench}, \emph{MMStar}~\citep{mmstar}, and \emph{CV-Bench-2D} and \emph{CV-Bench-3D}~\citep{cvbench}---spanning fine-grained visual search, high-resolution understanding, holistic multimodal reasoning, and 2D/3D spatial reasoning, and an \emph{extended} suite of five---BLINK~\citep{blink}, ChartQA~\citep{chartqa}, CharXiv~\citep{charxiv}, PixmoCount~\citep{pixmo}, and MME-RealWorld~\citep{mmerealworld}. Headline claims are reported on the core suite; the extended suite is included for transparency. Concurrent agentic-TWI benchmarks such as TIR-Bench~\citep{tirbench} broaden tool-dependent task coverage; we focus on this perception-heavy core suite where the carrier-swap claim is most directly testable.

\paragraph{Comparison groups.} \textbf{Thinking-with-images:} standard training and inference with full image return. \textbf{\methodname{}:} call-but-no-return during both training and inference. \textbf{base:} unmodified Qwen2.5-VL-7B-Instruct, reported under Direct MC and Agent Mode for calibration.

\paragraph{Evaluation protocol.} Tool-trained models are evaluated in Agent Mode with multi-turn tool calling, using either image feedback or the \methodname{} text-only feedback counterpart; the Direct MC base row provides a single-turn baseline, while the Agent Mode base row shows the effect of the same runner without tool training. Each trained model is evaluated under the protocol it was trained to follow. Hardware, compute budget, and dataset licenses are reported in Appendix~\ref{sec:exp_setup}.

\subsection{\texorpdfstring{Main Results: \methodname{} $\geq$ thinking-with-images Across Scales}{Main Results: \methodnameplain{} >= thinking-with-images Across Scales}}
\label{sec:main_results}

The key carrier-swap comparisons in Table~\ref{tab:main_results} are 9.5K thinking-with-images vs.\ 9.5K \methodname{} and 65K thinking-with-images vs.\ 65K \methodname{}. The \methodname{} SFT+RL row reports post-training behavior, while the remaining rows provide external reference points and base-model context.

\paragraph{\methodname{} $\geq$ thinking-with-images across SFT scales.} At the 9.5K LoRA scale (LoRA used because cold-start full FT on 9.5K erodes the base model's agent format prior and mode-collapses), \methodname{} reaches \ResultMainGFourBypassSftNineKVstar{} vs.\ thinking-with-images at \ResultMainGThreeTwiSftNineKVstar{} (\ResultMainGapSftNineKVstar{}). At the 65K full fine-tuning scale, the V*Bench gap is \ResultMainGapSftSixtyFiveKVstar{}, and the core 6-bench mean gap between \methodname{} and thinking-with-images---each evaluated under the protocol it was trained to follow (\methodname{} text-only feedback vs.\ Agent Mode with image return)---is $+1.39$\,pp (four gains, two small reversals; largest reversal 1.05\,pp).

\paragraph{RL is not required for \methodname{} to match thinking-with-images.} The carrier-swap result is established at SFT alone: at the 65K SFT scale, \methodname{} already matches or exceeds thinking-with-images on the 6-bench mean (Table~\ref{tab:main_results}). Adding GRPO on top of the 65K \methodname{} SFT model is reported as post-training behavior rather than as an accuracy-gain claim. Under identical RL hyperparameters (batch 128, lr $1{\times}10^{-6}$, 64 GPUs), the thinking-with-images model abandons tool use entirely by step~300 and remains at 0\% tool-call rate at step~800, while the \methodname{} model recovers from a transient step-200 collapse and stabilizes at 75\% mean tool-call rate (Figure~\ref{fig:rl_dynamics} and Table~\ref{tab:rl_tool_rate} in Appendix~\ref{sec:rl_dynamics}). Stable thinking-with-images RL has been reported with a larger cold-start corpus that includes an unreleased Long-CoT component~\citep{deepeyesv2}; our results are consistent with---not contradicted by---that report. We do not claim a cross-system sample-efficiency result; we report only the internal \methodname{}-vs.-thinking-with-images contrast under matched 65K-no-Long-CoT conditions, where the carrier is the only variable. Full RL hyperparameters, the step-200 transient in \methodname{}, and per-step dynamics are reported in Appendix~\ref{sec:rl_dynamics}.

\paragraph{Auxiliary observation: thinking-with-images can underperform the base model on a spatial sub-category.} \emph{This paragraph reports a side effect of thinking-with-images training and is not part of the causal argument above.} At 65K full FT, thinking-with-images posts a positive macro-average gain on V*Bench Agent Mode (\ResultMainGThreeOverBaseSftSixtyFiveKVstar{} over the base \ResultMainBaseQwenAgentVstar{}), but the gain is unevenly distributed: on the V*Bench \emph{relative\_position} sub-category, accuracy drops from 71.05\% to 57.89\% ($-$13.16\,pp). Tool-call training can damage spatial-relation reasoning even when the macro average improves.

\begin{summarybox}
\methodname{} matches or exceeds thinking-with-images without post-call pixels under SFT, and maintains active tool use under RL where thinking-with-images collapses. The post-call pixel return is not a necessary carrier for tool-augmented visual reasoning gains.
\end{summarybox}

\section{Why \methodname{} Works: The Scaffold Carries the Gain}
\label{sec:mechanism}

Having shown that the pixel carrier can be removed without reliable loss (Section~\ref{sec:causal}), we now ask what remains when the image return is gone. Two complementary analyses address this: a paired audit that decomposes image and scaffold contributions in the training corpus, and a controlled decomposition experiment identifying which textual components of the scaffold are load-bearing.

\subsection{Data Audit: Scaffold Recovers Image Accuracy}
\label{sec:data_audit}

We audit DeepEyes V2 training trajectories by varying the information given to a Gemini-3-Flash judge (temperature 0). Each paired sample is drawn from the perception and computation subsets. The scaffold consists only of the first \texttt{<think>} reasoning before code execution and the crop/zoom code blocks; it excludes execution results, later reasoning, and the final answer. This makes the audit a mechanism probe for Section~\ref{sec:causal}, not an independent causal claim.

\begin{table}[t]
\centering
\small
\caption{\textbf{Factorial audit of DeepEyes V2 training trajectories.} Rows define the information visible to the Gemini-3-Flash judge for the same paired $N=1{,}000$ samples. \emph{Scaffold} denotes the structured spatial prompt extracted from the first assistant reasoning block and crop/zoom code; no tool execution result or final answer is included.}
\label{tab:factorial_audit}
\begin{tabular}{@{}llc@{}}
\toprule
Condition & Judge input & Accuracy \\
\midrule
Question only & Question & 51.20\% \\
Image only & Question + image & 73.50\% \\
Scaffold only & Question + scaffold, no image & 73.10\% \\
Image + scaffold & Question + image + scaffold & \textbf{79.40\%} \\
\bottomrule
\end{tabular}
\end{table}

Table~\ref{tab:factorial_audit} shows that the scaffold alone recovers accuracy equivalent to the image-only condition: 73.10\% vs.\ 73.50\% on $N=1{,}000$ paired samples. A paired bootstrap for scaffold-only minus image-only gives $-0.40$\,pp with 95\% CI $[-3.20,+2.40]$, passing a 5\,pp non-inferiority test. The combined condition reaches 79.4\%, consistent with the scaffold and image being complementary in a minority of examples rather than one strictly dominating the other.

The gain decomposition is similarly balanced. Relative to the question-only baseline, the image-plus-scaffold condition gains 28.2\,pp: 57\% is redundant accuracy gain recovered by both image and scaffold, 22\% is image-unique, and 21\% is scaffold-unique. At the per-sample level, image and scaffold are both correct on 62.8\% of examples, image-only on 10.7\%, scaffold-only on 10.3\%, and neither on 16.2\%.

This helps explain why the training-time \methodname{} intervention in Section~\ref{sec:causal} can work: for many DeepEyes V2 traces, the structured spatial prompt already carries the recoverable accuracy gain that would otherwise be attributed to the returned image. The pixel return remains useful in some cases, but it is not the dominant load-bearing carrier under this training distribution. Details of the audit protocol, including paired bootstrap uncertainty and gain decomposition, are in Appendix~\ref{sec:data_audit_details}.

\subsection{Scaffold Decomposition: Reasoning Guides and Code Grounds}
\label{sec:scaffold-decomposition}

The factorial audit identifies a redundant accuracy gain at the corpus level; it does not say which \emph{component} of the scaffold carries the active signal. Each assistant turn in a \methodname{} trajectory contains two distinct text elements: (1)~\emph{reasoning text} inside \texttt{<think>} blocks that explains \emph{why} a crop region is chosen, and (2)~\emph{code body} that specifies the spatial coordinates. We isolate their contributions by training three models on the same 9.5K \methodname{} data, varying only the text content of assistant messages:

\begin{itemize}
  \item \textbf{Full scaffold}: original training data with reasoning + executable code.
  \item \textbf{Reasoning only}: reasoning preserved, code body replaced with \texttt{pass}.
  \item \textbf{Code only}: reasoning emptied, executable code preserved.
\end{itemize}

\begin{figure}[t]
\centering
\includegraphics[width=\linewidth]{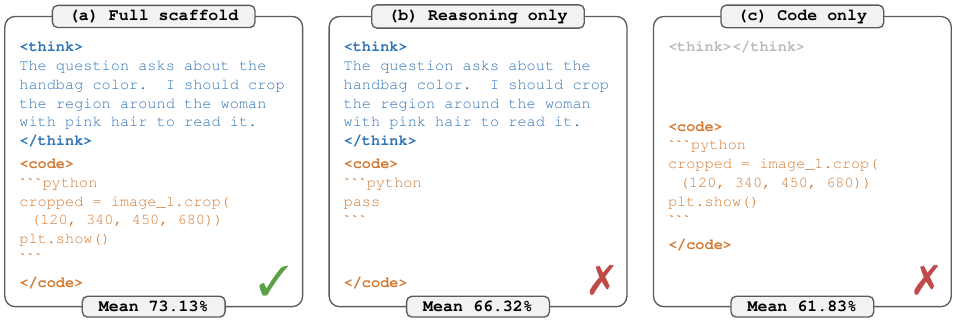}
\caption{\textbf{Scaffold component decomposition: assistant content.} The three 9.5K LoRA variants share identical training questions and tool-return policy; only the assistant text of each tool turn is manipulated. Reasoning supplies the \emph{why} (which region matters); code supplies the \emph{where} (concrete coordinates). Removing either component degrades mean accuracy across V*Bench, HR-4K, and CV-2D.}
\label{fig:scaffold_decomposition}
\end{figure}

All three are evaluated under Agent Mode with \methodname{} text-only feedback (the natural inference protocol for \methodname{}-trained models), using the same judge and pipeline as Table~\ref{tab:main_results}. Figure~\ref{fig:scaffold_decomposition} shows the manipulated assistant content schematically; Table~\ref{tab:degradation} reports the resulting per-benchmark accuracy.

\begin{table}[t]
\centering
\small
\caption{\textbf{Scaffold component decomposition.} All variants use the same 9.5K LoRA questions and Agent Mode evaluation with \methodname{} text-only feedback, judged by Gemini-3-Flash; only assistant text is manipulated. \emph{Full scaffold}=reasoning+code, \emph{Reasoning only}=reasoning+\texttt{pass}, \emph{Code only}=empty reasoning+code. Mean averages V*Bench ($N=191$), HR-4K ($N=800$), and CV-2D ($N=1{,}438$).}
\label{tab:degradation}
\begin{tabular}{@{}lcccr@{}}
\toprule
Variant & V* & HR-4K & CV-2D & Mean \\
\midrule
Full scaffold & \textbf{\ResultMainGFourBypassSftNineKVstarValue} & \ResultMainGFourBypassSftNineKHrBenchFourKValue & \textbf{\ResultMainGFourBypassSftNineKCvBenchTwoDValue} & \textbf{73.13} \\
Reasoning only & \ResultSecFourDegradationLTwoVstarValue & \textbf{\ResultSecFourDegradationLTwoHrBenchFourKValue} & \ResultSecFourDegradationLTwoCvBenchTwoDValue & 66.32 \\
Code only & \ResultSecFourDegradationLOneVstarValue & \ResultSecFourDegradationLOneHrBenchFourKValue & \ResultSecFourDegradationLOneCvBenchTwoDValue & 61.83 \\
\bottomrule
\end{tabular}
\end{table}

Table~\ref{tab:degradation} shows that removing either component degrades average \methodname{} performance. Removing reasoning while preserving code causes the largest drop ($-$11.30\,pp mean): the model continues to emit executable \texttt{.crop()} calls, but under \methodname{} these produce no visual feedback, resulting in \emph{hallucinated spatial attention}---the model acts as if it observed a region it never saw. Removing code while preserving reasoning also degrades performance on average ($-$6.81\,pp mean), though the effect is benchmark-dependent: HR-4K shows no degradation (reasoning only is $+0.62$\,pp above the full scaffold; reasoning alone suffices for high-resolution search), while V*Bench ($-$9.94\,pp) and CV-2D ($-$11.12\,pp) require code grounding to anchor spatial reasoning to concrete coordinates.

The ordering full scaffold $>$ reasoning only $>$ code only reveals a compositional structure: reasoning provides the \emph{why} (which region matters and what to look for), code provides the \emph{where} (specific coordinates that anchor the reasoning), and their combination forms the effective scaffold. Neither component alone recovers full performance. The full table (Table~\ref{tab:degradation_appendix}) including a skeleton-only baseline is in Appendix~\ref{sec:scaffold_decomp_details}.

Appendix~\ref{sec:jigsaw} provides a small Visual Jigsaw cross-check; we treat it as supporting evidence rather than an independent benchmark claim.

\begin{summarybox}
The scaffold that survives \methodname{} is a composition of reasoning text and spatial code. Reasoning supplies the task intent and visual rationale; code anchors it to concrete image coordinates. Removing either component weakens average performance, with code mattering most on V*Bench and CV-2D.
\end{summarybox}

\section{\methodname{} Is the Lower-Cost Default When Pixels Are Not Load-Bearing}
\label{sec:implications}

\begin{table}[t]
\centering
\small
\caption{\textbf{Cost--latency--accuracy trade-off at 65K cold-start.} Rows use Qwen2.5-VL-7B-Instruct and Agent Mode. The 6-bench mean averages V*Bench, HR-4K, HR-8K, MMStar, CV-2D, and CV-3D; --- marks latency-only rows without canonical accuracy. The \emph{API Calls} column counts returned-image tool executions per V*Bench sample. Latency is measured end-to-end on V*Bench ($N=191$, sequential, 2$\times$H20 TP=2). \methodname{} rows return a fixed text placeholder instead of pixels.}
\label{tab:tradeoff}
\resizebox{\textwidth}{!}{%
\begin{tabular}{@{}lccccccc@{}}
\toprule
Method & Eval & V*Bench & 6-bench mean & Turns & API Calls & Latency (mean) & Latency (p50) \\
\midrule
\methodname{} (sandbox)              & Agent & \ResultMainGFourBypassSftSixtyFiveKVstarValue{} & \textbf{71.23} & \textbf{2.3} & \textbf{0}   & \textbf{2.91\,s}  & \textbf{2.49\,s} \\
Thinking-with-images         & Agent & \textbf{\ResultMainGThreeTwiSftSixtyFiveKVstarValue{}} & 69.84 & 2.8 & 1.8 & 4.12\,s  & 3.10\,s \\
\methodname{} (no execution; latency only)\footnotemark     & Agent & --- & ---  & 2.4 & \textbf{0}   & 2.22\,s  & 1.73\,s \\
\bottomrule
\end{tabular}
}
\end{table}
\footnotetext{Latency-only reference row: tool calls are generated but not executed, so this row is not used for canonical accuracy comparisons. Canonical \methodname{} accuracy is reported by the sandbox row; no-execution latency represents the pure-LLM inference upper bound.}

The Tool-Call Scaffold Hypothesis yields two deployment-relevant consequences (Table~\ref{tab:tradeoff}).

\begin{figure}[t]
\centering
\includegraphics[width=\textwidth]{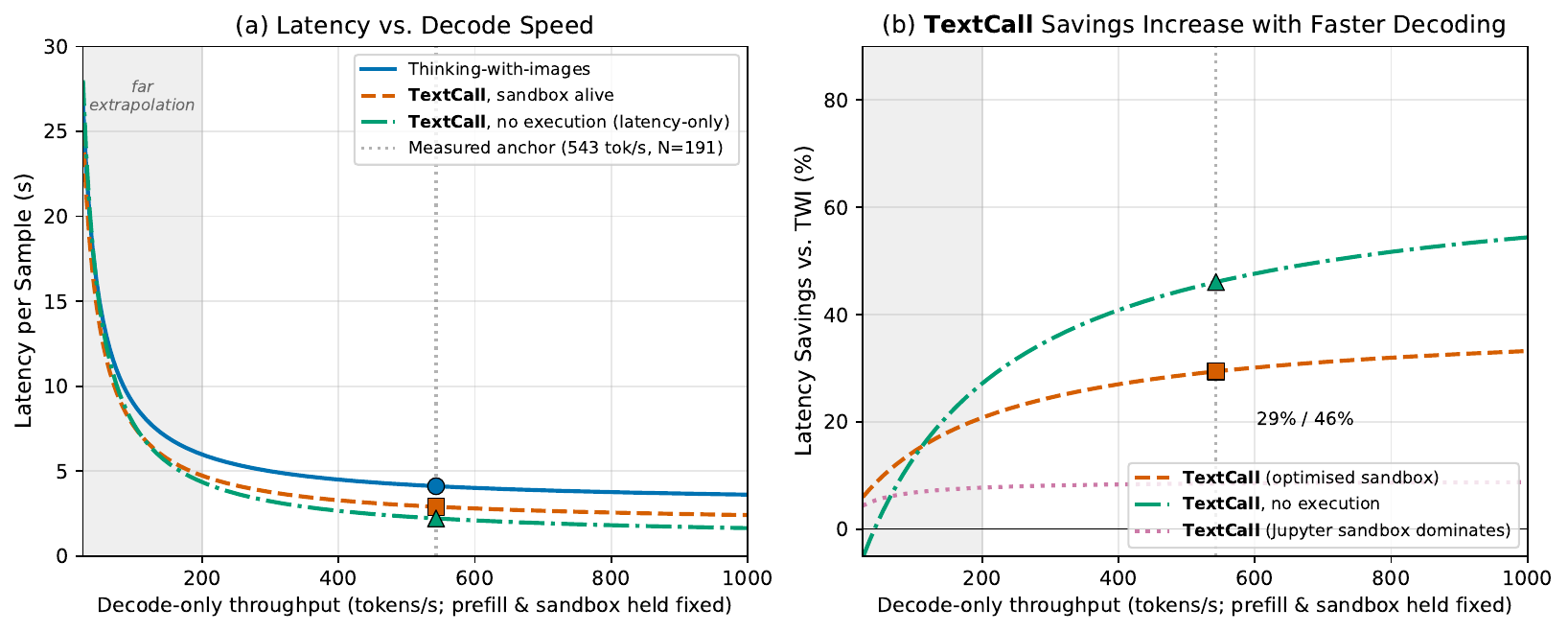}
\caption{\textbf{Token-speed sweep: \methodname{}'s latency advantage increases with faster decoding.} Curves vary decode speed while holding measured non-decode costs fixed. Markers show the measured V*Bench means at 543\,tok/s; the shaded region below 200\,tok/s marks far extrapolation where batching and scheduler effects are not modeled. (a)~Absolute latency per sample. \methodname{} avoids returned-image vision prefill, giving it a lower latency floor under our sequential per-sample implementation. (b)~Latency savings relative to thinking-with-images. The dotted line shows the original kernel-based sandbox setting, where a large shared sandbox cost reduces the savings at the measured anchor.}
\label{fig:speed_sweep}
\end{figure}

\paragraph{Within Agent~Mode, \methodname{} matches or exceeds thinking-with-images at lower deployment cost.}
Beyond preserving 6-bench accuracy ($+1.39$\,pp; \S\ref{sec:main_results}), \methodname{} eliminates image-returning tool executions and reduces end-to-end latency by 29--46\%.
The latency gap is architectural: each returned image incurs ${\sim}498$\,ms of vision-token prefill in subsequent turns, whereas sandbox execution contributes only 15\% of total latency.
Under the same decode-only sensitivity model used in Figure~\ref{fig:speed_sweep}, \methodname{}'s savings grow monotonically with decode speed: at the measured 543\,tok/s anchor, the canonical sandbox deployment saves $29\%$ and the latency-only floor saves $46\%$; doubling decode speed pushes the latency-only savings toward $\approx 50\%$, as the fixed per-image vision-prefill cost dominates thinking-with-images inference.

\paragraph{Deployment default.}
For tasks where the structured spatial prompt captures the task-relevant evidence, \methodname{} should be the default because it preserves accuracy while removing returned-pixel overhead.
Retain pixel returns when the task plausibly has a representational bottleneck or a visual prior gap, such as novel objects or fine-grained visual differences not well captured in text.
Meta-cognitive policies that allow the model to decline tool use are a complementary mitigation~\citep{actwisely}.
Extended boundary conditions under which pixel returns may become load-bearing are discussed in Appendix~\ref{app:boundary_extended}.

\section{Conclusion}
\label{sec:conclusion}

We posed the unit-of-thought question for tool-augmented visual reasoning and tested it with a training-time carrier swap: preserve the tool-call scaffold, remove the returned pixels, and evaluate the model under the protocol it was trained to follow.
Across the evaluated thinking-with-images regime, the results support the \emph{Tool-Call Scaffold Hypothesis}: much of the gain attributed to visual tool use is carried by the structured text emitted before image observation, not by the post-call pixel carrier.

This conclusion is deliberately scoped: one base model, one cold-start corpus, one seed, one GRPO epoch, and one perception-heavy evaluation suite.
The missing regime is therefore clear: tasks where returned pixels supply information the scaffold cannot verbalize or replace.
We did not find such a case on the suite-level mean, despite small per-benchmark thinking-with-images advantages (up to ${\sim}$1\,pp).
In the regime studied here, \methodname{} is a lower-cost default and future thinking-with-images systems should report a scaffold-only control so that the load-bearing carrier is measured rather than assumed.

\bibliographystyle{plainnat}
\bibliography{references}

\appendix
\section{Additional Experimental Setup}
\label{sec:exp_setup}

\paragraph{Hardware.}
All training and evaluation jobs use NVIDIA H20 GPUs. Per-run configurations are listed in Table~\ref{tab:compute}.

\paragraph{Compute usage.} Table~\ref{tab:compute} lists GPU configuration and approximate wall-clock per run for the experiments reported in the main paper. Wall-clock figures are rounded to reflect shared-cluster contention and exclude scheduler bring-up; the Agent Mode eval range spans V*Bench (191 samples, ${\sim}5$\,min on DP=4) at the low end and MME-RealWorld (23K samples, ${\sim}10$--12\,h including judge tail) at the high end. Totals are dominated by 65K full FT, RL post-training, and the canonical multi-benchmark Agent Mode evaluation sweep over MME-RealWorld.

\begin{table}[h]
\centering
\small
\caption{\textbf{Compute usage per reported run.} H20-class GPUs throughout; wall-clock values are approximate and exclude scheduler bring-up. $\dagger$ and $\ddagger$ are defined below the table.}
\label{tab:compute}
\begin{tabular}{@{}lll@{}}
\toprule
Run & Configuration & Wall-clock \\
\midrule
9.5K LoRA SFT (rank 8, per group) & $1\times8\times$H20 & $\sim$1 h \\
65K full FT SFT (cold-start, per group) & $8\times8\times$H20 & $\sim$13 h \\
RL post-training (GRPO, step 800)\textsuperscript{$\dagger$} & $8\times8\times$H20 & $\sim$25--30 h \\
Visual Jigsaw probe RL (GRPO, $n=50$) & $1\times8\times$H20 & $\sim$8 h \\
Agent Mode eval (per ckpt $\times$ bench, DP=4)\textsuperscript{$\ddagger$} & $1\times8\times$H20 + judge & $\sim$0.1--12 h \\
\bottomrule
\end{tabular}

\vspace{2pt}
{\footnotesize \textsuperscript{$\dagger$}\,${\sim}1$ epoch, batch 128, lr $1{\times}10^{-6}$. \textsuperscript{$\ddagger$}\,SGLang~\citep{sglang} runner.}
\end{table}

\paragraph{Main SFT and evaluation setup.} The main 65K thinking-with-images/\methodname{} SFT runs follow the public DeepEyesV2 cold-start configuration unchanged so that the only manipulated variable is the tool-return path. The 9.5K runs use LoRA fine-tuning (rank 8, learning rate $10^{-4}$, 3 epochs), matching the small-scale protocol used for the corresponding ablations. Tool-trained models are evaluated in Agent Mode with either image feedback or \methodname{} text-only feedback. The base model is reported under Direct MC as a single-turn baseline and under Agent Mode for runner calibration. All Agent/\methodname{} evaluations use the same VLMEvalKit-canonical runner and judge configuration across compared checkpoints.

\paragraph{Existing assets and licenses.} Table~\ref{tab:licenses} lists the third-party assets used in this work and their licenses or terms of use. All assets are used within their stated terms.

\begin{table}[h]
\centering
\small
\caption{\textbf{Existing assets used in this work.}}
\label{tab:licenses}
\begin{tabular}{@{}lll@{}}
\toprule
Asset & Type & License / Terms \\
\midrule
Qwen2.5-VL-7B-Instruct~\citep{qwen25vl} & Base model & Apache-2.0 \\
DeepEyesV2 cold-start corpus~\citep{deepeyesv2} & Training data (SFT) & Research-only, per official release \\
DeepEyesV2 RL corpus~\citep{deepeyesv2} & Training data (RL) & Research-only, per official release \\
V*Bench~\citep{vstar} & Benchmark & MIT, per official release \\
HR-Bench-4K / 8K~\citep{hrbench} & Benchmark & Apache-2.0, per official release \\
MMStar~\citep{mmstar} & Benchmark & Apache-2.0, per official release \\
CV-Bench (2D / 3D)~\citep{cvbench} & Benchmark & MIT, per official release \\
MME-RealWorld~\citep{mmerealworld} & Benchmark & Research-only, per official release \\
BLINK~\citep{blink} & Benchmark & CC-BY-4.0, per official release \\
CharXiv~\citep{charxiv} & Benchmark & MIT, per official release \\
ChartQA~\citep{chartqa} & Benchmark & GPL-3.0, per official release \\
PixmoCount~\citep{pixmo} & Benchmark & Apache-2.0, per official release \\
Zebra-CoT (Visual Jigsaw subset)~\citep{li2025zebra} & Benchmark / data & Research-only, per official release \\
verl~\citep{verl} & Training framework & Apache-2.0 \\
LlamaFactory~\citep{llamafactory} & Training framework & Apache-2.0 \\
VLMEvalKit~\citep{vlmevalkit} & Eval framework & Apache-2.0 \\
Gemini~3~Pro / Nano~Banana~Pro~\citep{gemini} & API (data synthesis) & Per Google ToS \\
\bottomrule
\end{tabular}
\end{table}

\section{Data Audit Details}
\label{sec:data_audit_details}

The headline statistics in Section~\ref{sec:mechanism} come from a four-condition factorial audit over all $N=1{,}000$ sampled DeepEyesV2 SFT trajectories (500 perception + 500 computation, seed 42). A Gemini-3-Flash judge (temperature 0) answers the same question under four input conditions: question only, question plus original image, question plus structured spatial prompt without image, and question plus both image and prompt.

\paragraph{Trajectory parsing.} Each trajectory is parsed into the user question, the original image, the first assistant reasoning block before code execution, the code blocks that specify crop/zoom operations, and the final answer. The structured spatial prompt used in the audit is restricted to the first \texttt{<think>} block and code blocks. It excludes tool execution results, later reasoning, and the final answer. We verified that the final-answer literal appears in fewer than 1\% of scaffold code blocks (5 out of 1{,}000 samples), indicating that direct answer copying is not driving the scaffold-only result.

\paragraph{Judge conditions.} The \emph{question-only} condition gives the judge only the question. The \emph{image-only} condition adds the original image. The \emph{scaffold-only} condition adds the structured spatial prompt but withholds the image. The \emph{combined} condition gives both the image and the prompt and serves as the reference. The resulting accuracies are 51.20\%, 73.50\%, 73.10\%, and 79.40\%, respectively.

\paragraph{Paired uncertainty.} We compare \emph{scaffold-only} against \emph{image-only} with paired bootstrap resampling over the same $N=1{,}000$ samples. The difference is $-0.40$\,pp with 95\% CI $[-3.20,+2.40]$, which passes a 5\,pp non-inferiority margin. Thus the structured spatial prompt recovers equivalent accuracy to the original image under this audit protocol.

\paragraph{Gain decomposition.} Relative to the \emph{question-only} baseline, the \emph{combined} condition gains 28.2\,pp. Of this total gain, 57\% is redundant accuracy gain recovered by both image and scaffold, 22\% is image-unique, and 21\% is scaffold-unique. The paired overlap is: both correct 62.8\%, image-only 10.7\%, scaffold-only 10.3\%, and neither 16.2\%.

\paragraph{Caveats.} The audit is descriptive mechanism evidence, not the primary causal claim. The causal claim is established by the training-time intervention in Section~\ref{sec:main_results}; this audit explains why that intervention can preserve accuracy under the evaluated training distribution.

\section{\methodname{} Scaffold-Decomposition Details}
\label{sec:scaffold_decomp_details}

Section~\ref{sec:scaffold-decomposition} decomposes the \methodname{} scaffold while keeping the multi-turn return policy fixed. All variants use the same 9.5K \methodname{} LoRA setting (rank 8, learning rate $10^{-4}$, 3 epochs) and the same Agent Mode evaluation protocol with \methodname{} text-only feedback and Gemini-3-Flash judging. The only manipulated field is the assistant text inside each trajectory.

\begin{table}[h]
\centering
\small
\caption{\textbf{Scaffold-decomposition variants.} All rows use the same 9.5K LoRA questions and Agent Mode evaluation with \methodname{} text-only feedback. The first three rows match Section~\ref{sec:scaffold-decomposition}; \emph{Skeleton only} keeps the multi-turn wrapper but removes both reasoning text and executable code.}
\label{tab:degradation_appendix}
\begin{tabular}{@{}llcccc@{}}
\toprule
Variant & Assistant content & V* & HR-4K & CV-2D & Mean \\
\midrule
Full scaffold & reasoning + executable code & \ResultMainGFourBypassSftNineKVstarValue{} & \ResultMainGFourBypassSftNineKHrBenchFourKValue{} & \textbf{\ResultMainGFourBypassSftNineKCvBenchTwoDValue{}} & \textbf{73.13} \\
Reasoning only & reasoning + \texttt{pass} & \ResultSecFourDegradationLTwoVstarValue{} & \ResultSecFourDegradationLTwoHrBenchFourKValue{} & \ResultSecFourDegradationLTwoCvBenchTwoDValue{} & 66.32 \\
Code only & empty reasoning + executable code & \ResultSecFourDegradationLOneVstarValue{} & \ResultSecFourDegradationLOneHrBenchFourKValue{} & \ResultSecFourDegradationLOneCvBenchTwoDValue{} & 61.83 \\
Skeleton only & empty reasoning + \texttt{pass} & \textbf{\ResultAppDegradationLThreeVstarValue{}} & \textbf{\ResultAppDegradationLThreeHrBenchFourKValue{}} & \ResultAppDegradationLThreeCvBenchTwoDValue{} & 71.63 \\
\bottomrule
\end{tabular}
\end{table}

The decomposition supports a benchmark-dependent reading. Reasoning alone is sufficient on HR-4K but loses spatial grounding on V*Bench and CV-Bench-2D, while code without reasoning produces hallucinated spatial attention under the no-pixel-return protocol. The skeleton-only variant shows that the multi-turn wrapper itself is benign: because both reasoning and code are empty, the model learns no scaffold content and effectively falls back to the base model's direct-answering capability within the agent format. Its high accuracy therefore reflects base-model strength, not scaffold contribution, and should not be read as evidence that structure alone carries the gain.

\section{\methodname{} Return-Text Ablation}
\label{sec:method_return_ablation}

Our \methodname{} intervention replaces every returned image with the literal text \texttt{[Image output skipped]} and changes nothing else in the trajectory. This fixed-sentinel return policy keeps the model's tool-call scaffold intact (the model still produces \texttt{crop}/\texttt{zoom} arguments, code, search queries, etc.) while removing the post-call pixel carrier.

\paragraph{No caption or oracle information enters training.} All \methodname{} runs reported in Table~\ref{tab:main_results} (\methodname{} at 9.5K LoRA, \methodname{} at 65K full FT, \methodname{} SFT+RL) use the same fixed placeholder: 100\% of samples replace the returned image with \texttt{[Image output skipped]}. We never provide captions of returned images or ground-truth-derived answer hints, so the reported carrier-swap results isolate the fixed text sentinel from caption-return or oracle-return alternatives.

\paragraph{What this ablation does \emph{not} address.} The fixed-placeholder intervention cannot tell us whether a higher-quality external caption would have helped the model, nor whether oracle hints would have closed the residual gaps on benchmarks where neither thinking-with-images nor \methodname{} reaches state-of-the-art. We treat such positive-evidence experiments as future work, consistent with the limitations discussed in Section~\ref{sec:conclusion}.

\paragraph{Execution policies and latency measurement.} The canonical \methodname{} accuracy rows execute the same agent loop as thinking-with-images, but the observation returned to the model is always the fixed text placeholder rather than a new image. The latency-only no-execution row in Table~\ref{tab:tradeoff} is a separate reference: the model still emits tool-call text, but the sandbox call is skipped after generation. This row estimates the pure inference lower bound and is not used for accuracy comparisons.

\section{Qualitative Scaffold Examples}
\label{sec:scaffold-gallery}

\begin{figure}[!b]
    \centering
    \includegraphics[width=0.82\textwidth]{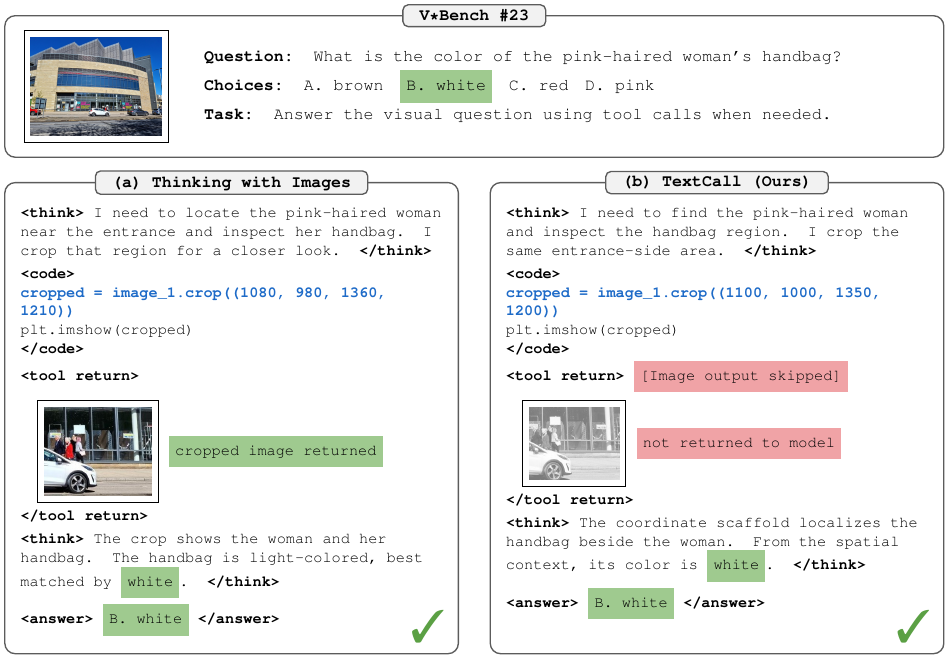}
    \caption{\textbf{Qualitative example A (V*Bench \#23, direct\_attributes).} ``What is the color of the pink-haired woman's handbag?'' The thinking-with-images model crops the region around the woman and reads the handbag color from the returned image; \methodname{} issues a closely matched crop but receives only a text placeholder, yet still correctly identifies the white handbag from the spatial scaffold context.}
    \label{fig:scaffold-gallery-a}
\end{figure}

\begin{figure}[p]
    \centering
    \includegraphics[width=0.92\textwidth]{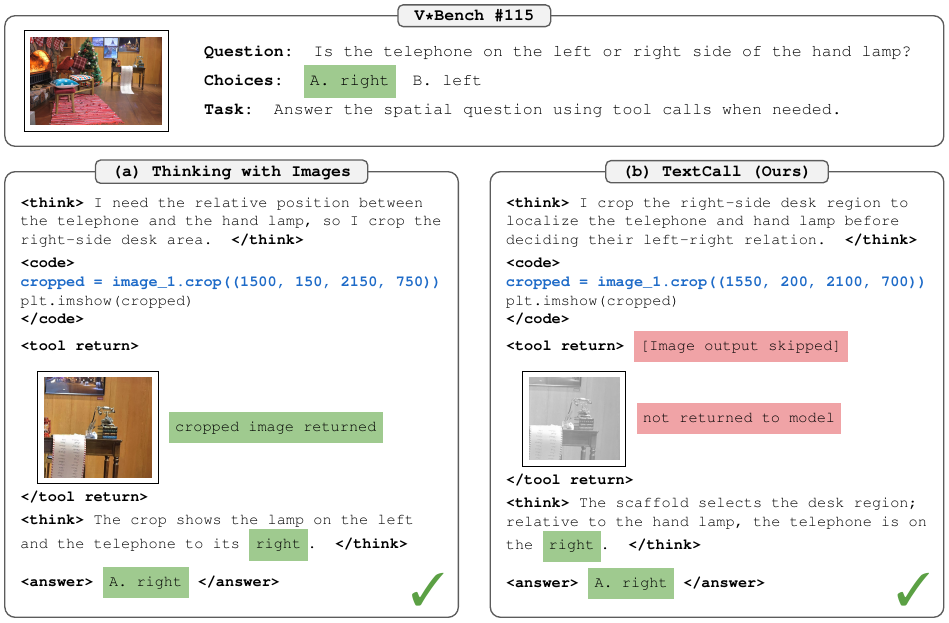}
    \caption{\textbf{Qualitative example B (V*Bench \#115, relative\_position).} ``Is the telephone on the left or right side of the hand lamp?'' A spatial-relation question requiring precise localization. Both paradigms crop the desk area; \methodname{} infers the relative position from the coordinate scaffold without seeing the cropped pixels.}
    \label{fig:scaffold-gallery-b}
\end{figure}

This appendix complements the main-paper scaffold overview (Figure~\ref{fig:scaffold-overview}) with additional V*Bench cases where the two Agent Mode paradigms, thinking-with-images and \methodname{}, answer correctly. Each case uses the same two-column layout as the main figure: the left panel shows the thinking-with-images model's multi-turn interaction with a real cropped image returned, while the right panel shows \methodname{}'s closely matched tool-call scaffold with the image return replaced by a text placeholder. These qualitative cases illustrate the carrier-swap behavior behind the controlled experiments. Figures~\ref{fig:scaffold-gallery-a} and~\ref{fig:scaffold-gallery-b} present two additional cases spanning color and spatial-relation queries.

\section{Visual Jigsaw Probe: Cross-Tool-Family Consistency Check}
\label{sec:jigsaw}

The main experiments (Section~\ref{sec:causal}) use crop/zoom/code tools. To check whether the no-degradation pattern extends to a qualitatively different tool family, we construct a small \emph{Visual Jigsaw} probe ($n=50$) using a text-to-image (T2I) generator. The probe is a consistency check, not a benchmark: it uses 50 manually verified samples filtered only for annotation validity, never on model outcomes.

\providecommand{\jigsawfirstpagewidth}{\textwidth}
\providecommand{\jigsawsecondpagewidth}{\textwidth}

\begin{figure}[!t]
\centering
\includegraphics[width=\jigsawfirstpagewidth]{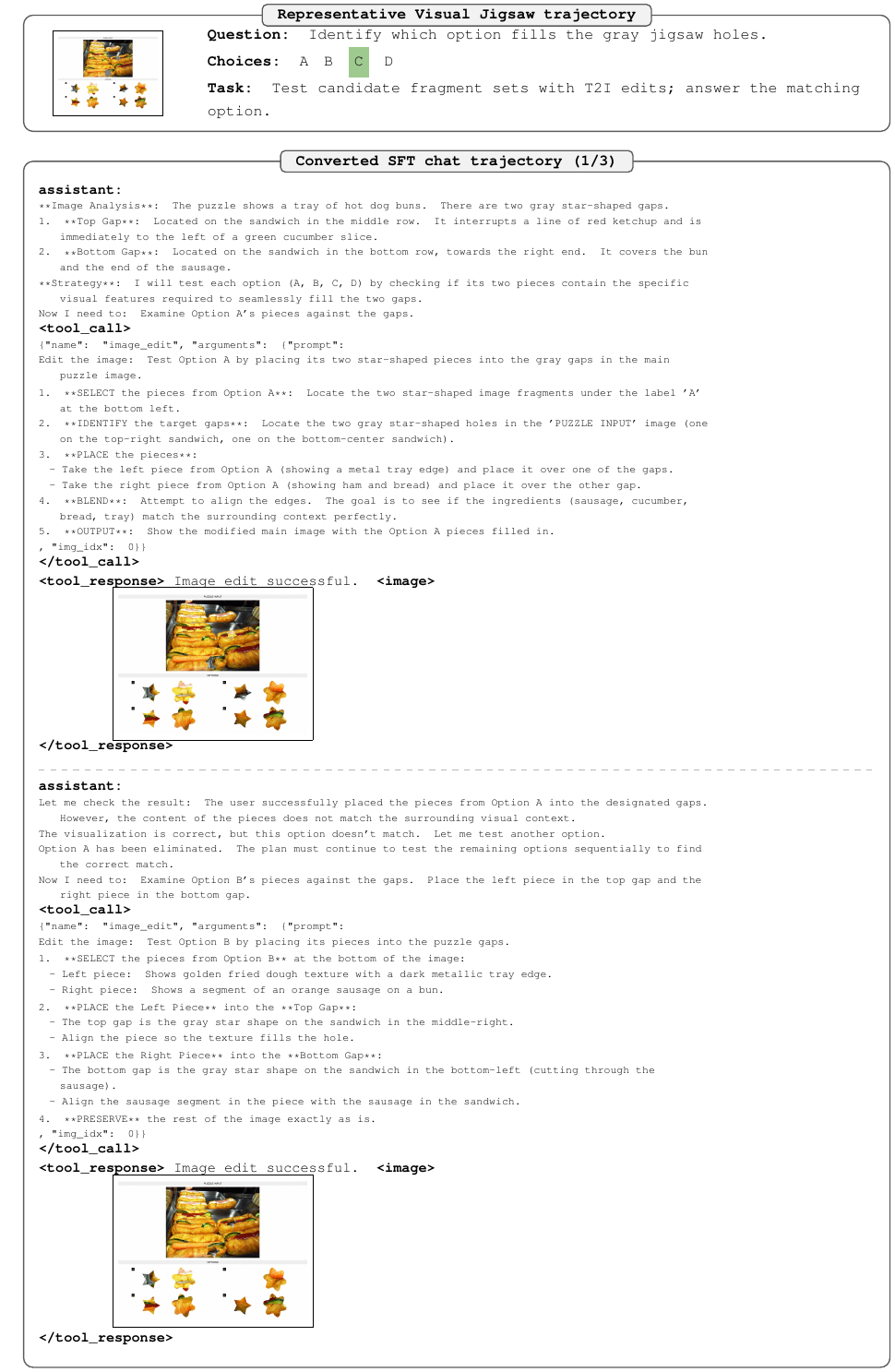}
\caption{\textbf{Visual Jigsaw probe: converted SFT chat trajectory.} The figure follows the post-conversion chat format: Agent messages interleave raw \texttt{image\_edit} tool calls with returned images while testing candidate options before committing to Option~C.}
\label{fig:jigsaw-example}
\end{figure}

\begin{figure}[!t]
\ContinuedFloat
\centering
\includegraphics[width=\jigsawsecondpagewidth]{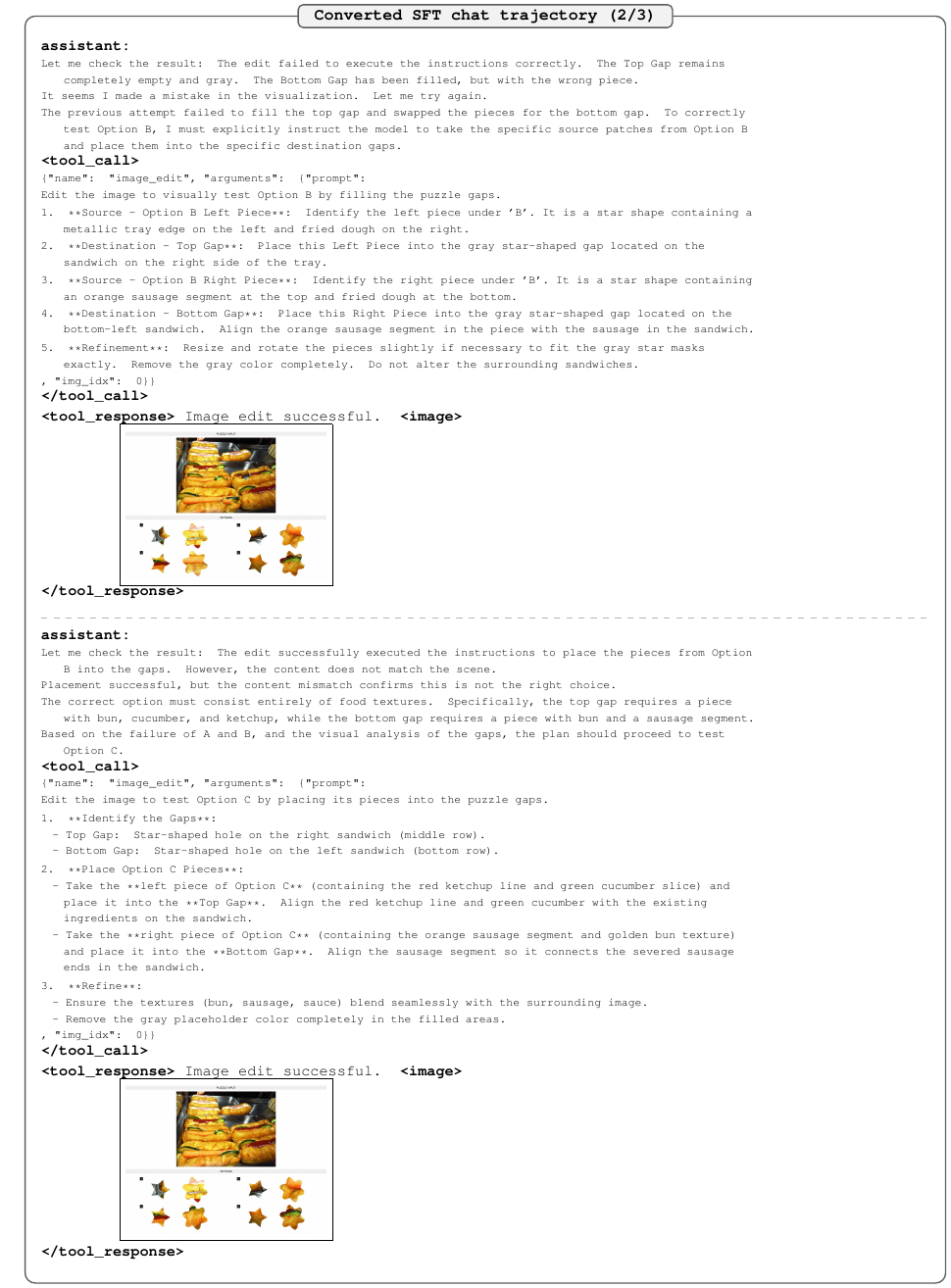}
\caption{\textbf{Visual Jigsaw probe: converted SFT chat trajectory, continued.} The second page shows the retry of Option~B after the previous tool response failed to fill the top gap.}
\end{figure}
\clearpage

\paragraph{Design.} The probe is built on the \emph{Visual Jigsaw} subset of Zebra-CoT~\citep{li2025zebra}. We synthesize paired SFT corpora with a Planner--Actor--Critic--Replanner multi-agent pipeline implemented in LangGraph, following the plan-and-solve and reflective-critic patterns established in prior agent work~\citep{wang2023planandsolve,react,shinn2023reflexion}: a Gemini~3~Pro \emph{planner} decomposes each Zebra-CoT ground-truth solution into a structured step list; a Gemini~3~Pro \emph{actor} rewrites the current step into a text-to-image instruction; Nano~Banana~Pro executes the image edit; a Gemini~3~Pro \emph{critic} verifies whether the returned image matches the expected sub-goal; and a Gemini~3~Pro \emph{replanner} decides whether to accept, retry the current step, or revise downstream steps based on the critic's verdict. The two corpora differ only in the tool-return slot: one keeps the T2I image, the other replaces it with \texttt{``Image edit successful. [no image returned]''}. All other tokens are byte-identical. Figure~\ref{fig:jigsaw-example} illustrates the converted SFT chat trajectory after this pipeline: Agent outputs, tool calls, and tool responses appear in message order, with the raw scaffold content preserved inside each tool-call prompt.

Both runs use the same base (Qwen2.5-VL-7B-Instruct), SFT config (LoRA rank-8, 10 epochs), and RL config (GRPO via verl, $8\times$H20). The only manipulated variable is the tool-return path. The real-tool run is evaluated under two policies: \emph{with-tools} (T2I image returned) and \emph{no-tool} (image suppressed at eval).

\begin{figure}[!t]
\ContinuedFloat
\centering
\includegraphics[width=\textwidth]{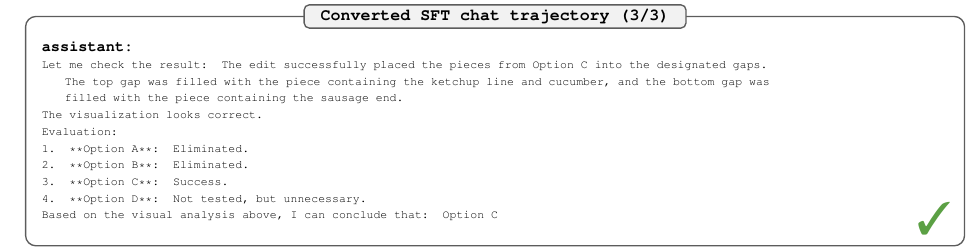}
\caption{\textbf{Visual Jigsaw probe: converted SFT chat trajectory, continued.} All four tool-call blocks preserve the raw \texttt{image\_edit} prompts from the training JSON. Figure~\ref{fig:jigsaw-levels} decomposes such prompts into structure, spatial grounding, and content description; Table~\ref{tab:jigsaw-results} reports the placeholder-return accuracy.}
\end{figure}

\paragraph{Results.} We retain RL checkpoints with tool-execution failure $<40\%$. Trajectory-mean accuracy across retained checkpoints is indistinguishable: $41.1\%$ (placeholder return) vs.\ $41.2\%$ (real-tool with-tools) and $44.6\%$ (real-tool no-tool). Table~\ref{tab:jigsaw-results} reports per-run peaks with paired-bootstrap CIs.

\begin{table}[h]
\centering
\small
\caption{\textbf{Visual Jigsaw probe results ($n=50$, paired).} Traj.\ mean averages retained RL checkpoints; Peak acc.\ reports best checkpoint accuracy and step. $\Delta$ is computed against the placeholder-return peak with paired bootstrap $B=10{,}000$; --- marks the reference row.}
\label{tab:jigsaw-results}
\begin{tabular}{lccc}
\toprule
Run (eval return policy) & Traj.\ mean & Peak acc.\ (step) & $\Delta$ vs.\ placeholder (95\% CI, $p$) \\
\midrule
Placeholder return & $41.1\%$ & \textbf{33/50=66\%} (s120) & --- \\
Real-tool (with-tools) & $41.2\%$ & $26/50=52\%$ (s50/90) & $+14$pp $[-4,+32]$, $p=0.13$ \\
Real-tool (no-tool) & \textbf{44.6\%} & $27/50=54\%$ (s100) & $+12$pp $[-6,+30]$, $p=0.23$ \\
\bottomrule
\end{tabular}
\end{table}

\paragraph{Interpretation.} The trajectory-mean parity is consistent with the no-degradation pattern observed in the main crop/zoom experiments. The wide bootstrap CIs reflect the small sample size ($n=50$) and preclude strong per-checkpoint claims; the relevant signal is that parity holds across retained checkpoints rather than only at a selected peak. We therefore treat the probe as cross-tool-family consistency evidence, not as an independent benchmark claim.

\begin{figure}[h]
\centering
\includegraphics[width=\textwidth]{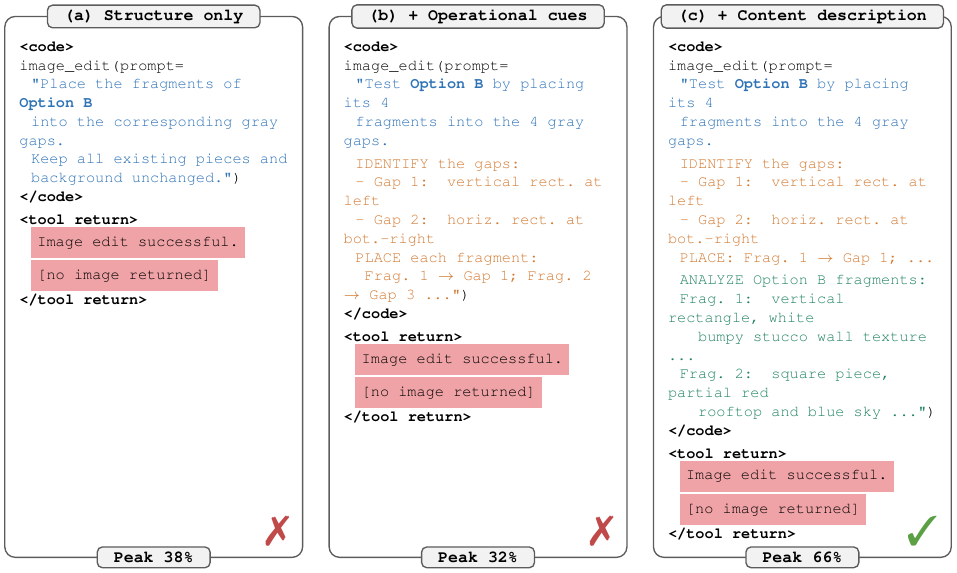}
\caption{\textbf{Jigsaw scaffold content levels.} Three levels of tool-call prompt content used in the scaffold ablation, sharing the same fixed placeholder return. \emph{Structure only} retains only the structural call (option ID and action verb); \emph{+ Operational cues} adds gap labels and fragment-to-gap mapping; \emph{+ Content description} further adds fragment textures. Peak placeholder-return accuracies are 38\% (structure only), 32\% (+ operational cues), and 66\% (+ content description); see Section~\ref{sec:jigsaw} for paired-bootstrap analysis.}
\label{fig:jigsaw-levels}
\end{figure}

\paragraph{Scaffold content ablation on Jigsaw.} Mirroring the decomposition in Section~\ref{sec:scaffold-decomposition}, we also varied the scaffold content within the T2I probe. Figure~\ref{fig:jigsaw-levels} shows the three levels, which differ only in how much textual information the tool-call prompt retains. The tool response is always \texttt{``Image edit successful. [no image returned]''} across all three levels, so only the pre-return scaffold content varies.

Peak placeholder-return accuracies are 38\% (structure only), 32\% (+ operational cues), and 66\% (+ content description). The content-bearing prompt provides the largest gain, consistent with the crop/zoom decomposition that textual scaffold content matters. The structure-only and + operational cues variants are statistically indistinguishable given the small sample size ($n=50$; paired bootstrap CIs overlap). Only + operational cues vs.\ real-tool thinking-with-images reaches significance ($\Delta=-20$\,pp, 95\% CI $[-38,-2]$, $p=0.04$). We report this as directional consistency evidence rather than a standalone finding.

\section{Extended Boundary Discussion}
\label{app:boundary_extended}

As a hypothesis-level extension to the limitations in Section~\ref{sec:conclusion}, two candidate conditions may make returned pixels load-bearing: (i) a \emph{representational bottleneck} where the visual evidence cannot be losslessly verbalized, and (ii) a \emph{prior-knowledge gap} concentrated in visual pretraining and poorly represented in text corpora. The benchmarks in our suite do not appear to jointly stress both conditions strongly. We expand here on candidate task families where these conditions might become active; verifying whether thinking-with-images then outperforms \methodname{} on such tasks remains future work.

\paragraph{Tasks requiring visual simulation.}~\citet{visgenunlocks} demonstrate that visual generation can unlock spatial reasoning that text-only training cannot, particularly for tasks requiring the model to mentally simulate physical transformations (e.g., rotating, folding, or intersecting objects). These tasks may satisfy both conditions: the spatial relationships are hard to describe verbally (bottleneck), and the dynamics are learned from visual experience (knowledge gap).

\paragraph{Tasks requiring fine-grained visual comparison.} Design and construction tasks require the model to compare subtle visual differences across intermediate states. If the differences exceed the granularity of natural language---e.g., slight curve variations in architectural drawings, or minor alignment offsets in mechanical assemblies---both conditions could be met: the differences are hard to verbalize (bottleneck) and the comparison skill is rooted in visual experience (knowledge gap).

We emphasize that these are \emph{hypotheses}, not conclusions. Verifying whether generated images provide genuine causal benefits on such tasks---using the same training-time intervention methodology---is a necessary next step. Our \methodname{} framework provides the experimental tool: if \methodname{} training matches thinking-with-images on these candidate tasks, returned pixels are still not empirically load-bearing under the tested protocol; if thinking-with-images significantly outperforms \methodname{}, the result would indicate that at least one candidate condition is active and should be analyzed directly. See Section~\ref{sec:implications} for the corresponding deployment recommendations.

\section{Diagnostic Evidence in the Literature}
\label{app:diagnostic_table}

Table~\ref{tab:diagnostic} consolidates prior evidence on the role of returned images in thinking-with-images pipelines, complementing the prose summary in Section~\ref{sec:related}. The picture is mixed: some studies find that returned pixels are under-used or redundant, while others show genuine benefits on perception-heavy tasks. Our contribution is the matched training-time carrier-swap counterpart that isolates the textual scaffold's role.

\begin{table}[h]
\centering
\footnotesize
\caption{\textbf{Diagnostic evidence about returned images in thinking-with-images systems.} Rows group prior inference-time, trace-audit, and training/data-side diagnostics; the final row denotes the matched training-time carrier-swap setting evaluated in this work.}
\label{tab:diagnostic}
\begin{tabular}{@{}llp{0.42\textwidth}@{}}
\toprule
\textbf{Paradigm} & \textbf{Paper} & \textbf{Key Finding} \\
\midrule
Visual-mark prompting & Set-of-Mark~\citep{setofmark} & Symbolic mark IDs enable textual spatial reference \\
Visual-CoT oracle probing & MIRA~\citep{mira} & Oracle visual cues yield $+$33.7\% gain, but models cannot self-generate them \\
End-to-end agentic RL & DeepEyes~\citep{deepeyes} & iMCoT (crop$+$text) outperforms text-only CoT; largest gap on high-res ($+$11.8pp HR-8K) \\
Causal intervention & Faithfulness~\citep{faithfulness} & Image$\to$noise: $-0.52\%$; text corrupted: $-13.09\%$ \\
Blind testing & Deltas~\citep{deltas} & Removing images: $+3.5\%$ \\
Statistical decomposition & MED~\citep{med} & ${>}$70\% gains from intrinsic reasoning \\
Faithfulness evaluation & CodeV~\citep{codev} & 57\% of crops contain the target \\
Attention reward & SAYO~\citep{sayo} & Original-image attention reward recovers most gains \\
Distillation & ZwZ~\citep{zwz} & Zoom benefits distilled into single-pass inference \\
Zero-vision SFT & Kimi K2.5~\citep{kimik25} & Visual reasoning can activate without visual data \\
Non-interleaved agentic CoT & sCoT~\citep{scot} & Textual reasoning trajectory (subagents see crops, return text only) outperforms pixel-interleaved iMCoT \\
\midrule
\textbf{Training-time ablation} & \textbf{Ours} & \textbf{\methodname{} $\geq$ thinking-with-images under aligned training/eval} \\
\bottomrule
\end{tabular}
\end{table}

\section{RL Training Dynamics: Thinking-with-images Collapse vs.\ \methodname{} Stability}
\label{sec:rl_dynamics}

We apply GRPO on top of both the matched thinking-with-images and \methodname{} 65K full-FT SFT checkpoints, using identical hyperparameters (batch 128, lr $1{\times}10^{-6}$, 64 GPUs, ${\sim}1$ epoch). The only difference is the tool-return policy during rollout: thinking-with-images returns full images; \methodname{} returns \texttt{[Image output skipped]}. We track tool-call rate (fraction of evaluation samples where the model emits at least one \texttt{<code>} block) across training steps.

\paragraph{Thinking-with-images collapses to direct answering.} Table~\ref{tab:rl_tool_rate} shows that the thinking-with-images model abandons tool use by step~300 and never recovers. At step~800 (and step~900, as a confirmation), tool-call rate is 0\% across all six core benchmarks. The model reverts to single-turn direct answering despite being evaluated in the full agent loop with available tools. Accuracy remains reasonable (e.g., V*Bench 75--80\%) because the base model's text reasoning is strong enough for direct answering on many samples---but the numbers no longer reflect tool-augmented exploration.

\paragraph{\methodname{} maintains active tool use.} The \methodname{} model shows a transient collapse at step~200 (100\% no-code on a diagnostic sweep) but recovers by step~400 and maintains 75\% mean tool-call rate at step~800 (Table~\ref{tab:rl_tool_rate}). The recovery suggests that, under this matched recipe, the scaffold-only return policy can preserve tool-use behavior at the reported checkpoint.

\paragraph{Mechanism interpretation.} The thinking-with-images collapse is consistent with the scaffold hypothesis, but it is not direct proof of the optimization pathway. One plausible interpretation is that, during RL exploration, the model finds a direct-answer policy with comparable reward to tool-augmented paths under this training distribution. RL can then favor the simpler policy with fewer turns and no tool overhead. In contrast, the \methodname{} checkpoint is trained and evaluated with the scaffold-only carrier, making active tool-call generation the protocol-aligned behavior at the reported checkpoint.

\paragraph{Comparison to \citet{deepeyesv2} and scope of our claim.} \citet{deepeyesv2} report stable thinking-with-images RL on top of a substantially larger cold-start corpus that includes an unreleased Long-CoT subset; the total cold-start scale and Long-CoT proportion are estimated from their Figure~9 and not stated explicitly in the paper. Our 65K SFT budget without Long-CoT is therefore not directly comparable to theirs. Beyond data scale, our setup also differs in RL algorithm (GRPO vs.\ DAPO), reward formulation, and the absence of prompt-level tool-benefit filtering. We do \emph{not} claim that our thinking-with-images collapse reproduces the pioneer-experiment failure mode of \citet{deepeyesv2} under matched conditions. We claim only an \emph{internally matched contrast}: under identical SFT data, identical RL recipe, identical reward, and identical training compute, the scaffold-only carrier survives RL at this budget while the pixel-return carrier does not. Stable thinking-with-images RL at higher SFT budgets is consistent with---not contradicted by---this finding.

\begin{figure}[h]
\centering
\includegraphics[width=\textwidth]{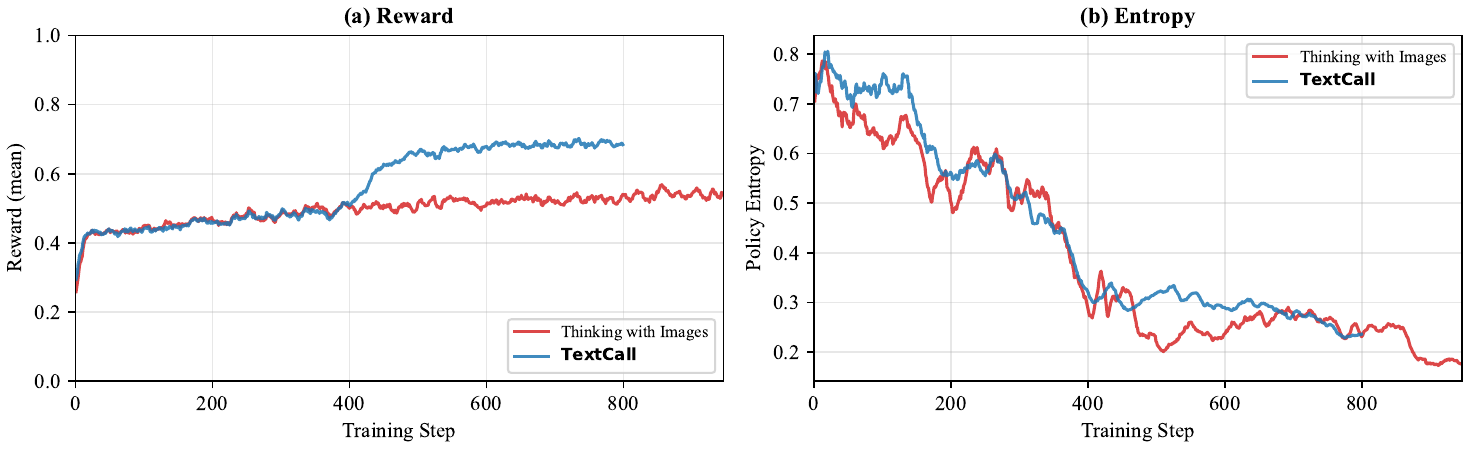}
\caption{\textbf{RL training dynamics from matched GRPO runs.} Curves show 10-step-smoothed training reward and policy entropy from the matched thinking-with-images and \methodname{} runs. These logged training metrics are not benchmark accuracy; benchmark-level tool-call rates are reported in Table~\ref{tab:rl_tool_rate}. The \methodname{} training log ends just before the evaluated step-800 checkpoint, while the step-800 checkpoint itself is included in the benchmark sweep.}
\label{fig:rl_dynamics}
\end{figure}

\begin{table}[h]
\centering
\small
\caption{\textbf{Tool-call rate (\%) across RL training steps.} Entries report the percentage of evaluation samples that emit at least one \texttt{<code>} block under Agent Mode. Step labels are checkpoint numbers; Mean averages the six listed benchmarks.}
\label{tab:rl_tool_rate}
\begin{tabular}{@{}lcccccc|c@{}}
\toprule
& V*Bench & HR-4K & HR-8K & MMStar & CV-2D & CV-3D & Mean \\
\midrule
\multicolumn{8}{@{}l}{\textit{Matched thinking-with-images RL}} \\
\quad Step 800  & 0.0 & 0.0 & 0.0 & 0.0 & 0.0 & 0.0 & 0.0 \\
\quad Step 900  & 0.0 & 0.0 & 0.0 & 0.5 & 0.0 & 0.0 & 0.1 \\
\midrule
\multicolumn{8}{@{}l}{\textit{Matched \methodname{} RL}} \\
\quad Step 700  & \textbf{86.9} & \textbf{79.6} & \textbf{75.8} & \textbf{72.5} & 85.7 & \textbf{100.0} & \textbf{83.4} \\
\quad Step 800  & 84.8 & 59.0 & 63.5 & 57.8 & \textbf{87.3} & 98.8 & 75.2 \\
\bottomrule
\end{tabular}
\end{table}

\paragraph{Why we report only \methodname{} RL in the main table.} The thinking-with-images step-800 numbers (V*Bench 75.4\%, HR-4K 76.6\%, etc.) reflect the base model's direct-answer reasoning under agent formatting, not tool-augmented exploration. Including them in a table designed to compare tool-use paradigms would be misleading. We therefore report only the \methodname{} RL result (step~800, 75\% tool-call rate) in Table~\ref{tab:main_results} and present the thinking-with-images dynamics here as supporting evidence for the scaffold hypothesis.

\paragraph{Caveats.} (i)~The step-200 transient collapse in \methodname{} means we cannot claim that \methodname{} is \emph{always} stable during RL---only that it recovers and stabilizes at the reported checkpoint. (ii)~The thinking-with-images collapse may partly reflect reward design: our reward ($0.8 \times \text{acc} + 0.2 \times \text{format}$) does not explicitly incentivize tool use. A tool-use bonus might delay the collapse, though prior work~\citep{deepeyesv2} reports that such bonuses trigger reward hacking. (iii)~These dynamics are specific to our training scale (7B model, 64 GPUs, ${\sim}1$ epoch); whether the pattern holds at larger scale remains open.

\section{Coordinate Precision Within the Code Body}
\label{sec:coord_precision}

Section~\ref{sec:scaffold-decomposition} shows that removing the code body degrades V*Bench by 9.94\,pp and CV-2D by 11.12\,pp, establishing that the syntactic structure of \texttt{.crop()} calls is load-bearing.
This leaves open whether the code component works by providing \emph{accurate numeric spatial evidence} or by imposing an \emph{explicit structured region-selection cue} regardless of coordinate accuracy.
We test this with a coordinate-precision ablation: matched models trained on correct versus shuffled coordinates.

\paragraph{Design.}
We render 3{,}317 verified training samples into matched arms differing only in the spatial evidence within the assistant reasoning:
\emph{correct coordinates} (real bounding-box values from verified object proposals),
\emph{shuffled coordinates} (values randomly permuted across samples---syntactically valid \texttt{.crop()} calls with spatially meaningless arguments),
and \emph{free CoT} (no coordinates or spatial markers; descriptive reference only, not matched to the Section~\ref{sec:scaffold-decomposition} code-removal ablation).
All arms train Qwen3-VL-8B-Instruct~\citep{qwen3vl} with LoRA rank~8 for 3 epochs and are evaluated under Direct MC think-mode with heuristic scoring on four benchmarks (HR-4K, MMStar, CV-2D, CV-3D).
Three seeds (42, 43, 44) are run for the primary correct-vs.-shuffled comparison.

\paragraph{Results.}
Table~\ref{tab:coord_precision} reports the cross-seed means.
Correct and shuffled coordinates are statistically equivalent: the item-level mean difference across all 14{,}806 paired predictions (3~seeds $\times$ 4~benchmarks) is $-0.11$\,pp (90\% CI $[-0.74,{+}0.51]$), passing a paired-bootstrap TOST equivalence test at $\delta{=}2$\,pp ($p{=}0.006$).
The equivalence margin is less than one-third of the 9.94\,pp code-removal effect in Section~\ref{sec:scaffold-decomposition}.
At the per-sample level, the two models produce identical predictions on 87.6\% of instances; on V*Bench (seed~42, 191~samples), concordance reaches 90.6\% with a symmetric discordant split (8~vs.~7, McNemar $p{=}1.0$).
A single-seed replication on the weaker Qwen2.5-VL-7B~\citep{qwen25vl} base is consistent with the same pattern (correct minus shuffled: $+0.13$\,pp).

\begin{table}[h]
\centering
\caption{\textbf{Coordinate precision ablation} (Qwen3-VL-8B, 3 seeds $\times$ 4 benchmarks, Direct MC think-mode). Mean $\pm$ std across seeds. TOST equivalence at $\delta{=}2$\,pp: $p{=}0.006$.}
\label{tab:coord_precision}
\small
\begin{tabular}{@{}lccccr@{}}
\toprule
Arm & HR-4K & MMStar & CV-2D & CV-3D & Mean \\
\midrule
Correct coordinates & $78.9{\scriptstyle\pm 0.5}$ & $65.1{\scriptstyle\pm 1.0}$ & $81.3{\scriptstyle\pm 0.6}$ & $88.0{\scriptstyle\pm 0.8}$ & 78.3 \\
Shuffled coordinates & $78.3{\scriptstyle\pm 1.1}$ & $66.2{\scriptstyle\pm 0.3}$ & $81.2{\scriptstyle\pm 0.4}$ & $88.8{\scriptstyle\pm 0.2}$ & 78.6 \\
Free CoT (no coords) & $78.4{\scriptstyle\pm 0.9}$ & $64.6{\scriptstyle\pm 0.5}$ & $81.3{\scriptstyle\pm 0.8}$ & $87.5{\scriptstyle\pm 0.4}$ & 77.9 \\
\midrule
Correct $-$ Shuffled & ${+}0.6$ & ${-}1.1$ & ${+}0.1$ & ${-}0.7$ & ${-}0.1$ \\
\bottomrule
\end{tabular}
\end{table}

\paragraph{Scale validation (9.5K, matched to Section~\ref{sec:scaffold-decomposition}).}
The 3K experiment above leaves open whether coordinate correctness might matter at larger data scales where the code-removal effect is established.
We therefore repeat the ablation on the full 9{,}500-sample \methodname{} training set used in Section~\ref{sec:scaffold-decomposition}, training Qwen2.5-VL-7B with the same LoRA configuration.
Coordinates in 2{,}193 samples containing \texttt{.crop()} calls are cross-sample permuted within coordinate type (pixel values shuffled with pixel values, ratios with ratios); the remaining 7{,}307 samples without \texttt{.crop()} are identical across arms.
Table~\ref{tab:coord_precision_9k} shows the same qualitative pattern: the mean difference is $-0.65$\,pp, far smaller than the ${\sim}10$\,pp code-removal effect from Section~\ref{sec:scaffold-decomposition}.

\begin{table}[h]
\centering
\caption{\textbf{Coordinate correctness at 9.5K scale} (Qwen2.5-VL-7B, seed 42, Direct MC think-mode, heuristic scoring). Same training recipe as the Section~\ref{sec:scaffold-decomposition} models.}
\label{tab:coord_precision_9k}
\small
\begin{tabular}{@{}lccccr@{}}
\toprule
Arm & HR-4K & MMStar & CV-2D & CV-3D & Mean \\
\midrule
Correct coordinates & 58.5 & 54.5 & 68.3 & 75.2 & 64.1 \\
Shuffled coordinates & 61.3 & 54.5 & 69.3 & 73.9 & 64.8 \\
\midrule
Correct $-$ Shuffled & ${-}2.8$ & ${0.0}$ & ${-}1.0$ & ${+}1.3$ & ${-}0.7$ \\
\bottomrule
\end{tabular}
\end{table}

\paragraph{Scope.}
The 3K multi-seed test establishes equivalence; the 9.5K matched-scale run shows the same qualitative pattern, arguing against a scale-specific artifact.
An important caveat: both experiments operate at the \emph{supervised fine-tuning} stage, where the base model has already acquired spatial understanding from vision--language pre-training.
Coordinate precision may well matter during \emph{pre-training}, where grounding data teaches the model what spatial coordinates mean in the first place (e.g., the bounding-box pre-training stages in Qwen2.5-VL~\citep{qwen25vl} and similar models).
Our finding is specific to the post-training regime: once a model already possesses spatial grounding from pre-training, the coordinate values in fine-tuning data serve as a structural cue rather than a source of new spatial knowledge.

\paragraph{Interpretation.}
Combined with Section~\ref{sec:scaffold-decomposition}, these results support a two-level decomposition: the \emph{format} of spatial code (having \texttt{.crop()} calls at all) matters substantially, while the \emph{numeric content} (whether coordinate values are instance-correct) does not detectably matter.
This pattern is consistent with the code body acting as a structured procedural cue that requires the model to specify a region before continuing its reasoning, rather than serving as a numerically accurate spatial reference.

\end{document}